\documentclass[10pt]{article}
\usepackage{thuidm}

\usepackage[utf8]{inputenc}
\usepackage{multirow}
\usepackage{colortbl}
\usepackage{tabularx}
\newcolumntype{Y}{>{\centering\arraybackslash}X}
\usepackage{wrapfig}
\usepackage{float}
\colorlet{uclablue}{thupurplemid}
\usepackage[ruled,vlined]{algorithm2e}
\SetKwInput{KwInput}{Input}
\SetKwInput{KwOutput}{Output}
\newtcolorbox{dataexample}[1]{
  breakable,
  colback=white,
  colframe=uclablue!75!black,
  colbacktitle=uclablue!12,
  coltitle=black,
  fonttitle=\bfseries,
  boxrule=0.6pt,
  arc=2pt,
  left=6pt,
  right=6pt,
  top=5pt,
  bottom=5pt,
  toptitle=2pt,
  bottomtitle=2pt,
  fontupper=\footnotesize,
  title={#1}
}

\usepackage{amsmath,amsfonts,bm}

\def\eqref#1{equation~\ref{#1}}

\def\1{\bm{1}}

\DeclareMathAlphabet{\mathsfit}{\encodingdefault}{\sfdefault}{m}{sl}
\SetMathAlphabet{\mathsfit}{bold}{\encodingdefault}{\sfdefault}{bx}{n}

\title{Tree-Aware Posterior Distillation for Long-Horizon\\Agentic Reinforcement Learning}

\author{%
  Heming~Zou\aff{1,2,*}, Qi~Wang\aff{1}, Yixiu~Mao\aff{1}, Yun~Qu\aff{1}, Yuyi~Guo\aff{1,2,*},
  Zeyuan~Liu\aff{1,2,*},\\
  Chenglin~Li\aff{1}, Weijie~Liu\aff{2}, Saiyong~Yang\aff{2}, Xiangyang~Ji\aff{1}%
}

\affiliation{%
  \aff{1}Tsinghua University,\quad
  \aff{2}Foundation Model Department, Tencent\\[-1pt]
  {\normalfont\fontsize{8.6}{10}\selectfont\aff{*}Work completed during an internship at Tencent.}%
}

\correspondence{%
  Qi~Wang (\href{mailto:cheemswang@mail.tsinghua.edu.cn}{cheemswang@mail.tsinghua.edu.cn})\\
  Saiyong~Yang (\href{mailto:stevesyang@tencent.com}{stevesyang@tencent.com})\\
  Xiangyang~Ji (\href{mailto:xyji@tsinghua.edu.cn}{xyji@tsinghua.edu.cn})%
}

\labwebsite[www.thuidm.com]{https://www.thuidm.com/}

\makeatletter
\titleformat{\section}
  {\sffamily\bfseries\fontsize{13.2}{15}\selectfont}
  {\textcolor{thupurpledark}{\thesection}}{0.75em}{\textcolor{thupurpledark}}
\titleformat{\subsection}
  {\sffamily\bfseries\fontsize{11.4}{13.5}\selectfont}
  {\textcolor{thupurpledark}{\thesubsection}}{0.7em}{\textcolor{thupurpledark}}
\titleformat{\subsubsection}
  {\sffamily\bfseries\fontsize{10.3}{12.5}\selectfont}
  {\textcolor{thupurpledark}{\thesubsubsection}}{0.65em}{\textcolor{thupurpledark}}
\titleformat{\paragraph}[runin]
  {\sffamily\bfseries\fontsize{9.8}{12}\selectfont}
  {}{0pt}{\textcolor{thupurpledark}}[\,]
\let\treeopsd@Needspace\Needspace
\newcommand{\treeopsd@skipNeedspace}{\nobreak\@ifstar{\@gobble}{\@gobble}}
\renewcommand{\Needspace}{%
  \if@nobreak\expandafter\treeopsd@skipNeedspace
  \else\expandafter\treeopsd@Needspace\fi}

\patchcmd{\maketitle}{\@title\par}{\centering\@title\relax\par}{}{\PackageWarning{main}{title patch 1 failed}}
\patchcmd{\maketitle}{\@title\par}{\centering\@title\relax\par}{}{\PackageWarning{main}{title patch 2 failed}}
\patchcmd{\maketitle}{\@author\par}{\centering\@author\relax\par}{}{\PackageWarning{main}{author patch 1 failed}}
\patchcmd{\maketitle}{\@author\par}{\centering\@author\relax\par}{}{\PackageWarning{main}{author patch 2 failed}}
\patchcmd{\maketitle}{\thuidm@affiliation\par}{\centering\bfseries\thuidm@affiliation\relax\par}{}{\PackageWarning{main}{affiliation patch 1 failed}}
\patchcmd{\maketitle}{\thuidm@affiliation\par}{\centering\bfseries\thuidm@affiliation\relax\par}{}{\PackageWarning{main}{affiliation patch 2 failed}}

\renewcommand{\thuidm@metaline}[2]{%
  \begingroup
  \fontsize{8.6}{11.4}\selectfont
  \sbox\z@{\sffamily\bfseries\color{thupurpledark}#1\hspace{0.35em}}%
  \hangindent\wd\z@ \hangafter\@ne
  \noindent\usebox\z@{\color{thuink}#2}\par
  \endgroup}

\newsavebox{\treeopsd@thulogo}
\newsavebox{\treeopsd@hylogo}
\sbox{\treeopsd@thulogo}{\includegraphics[height=9.5mm]{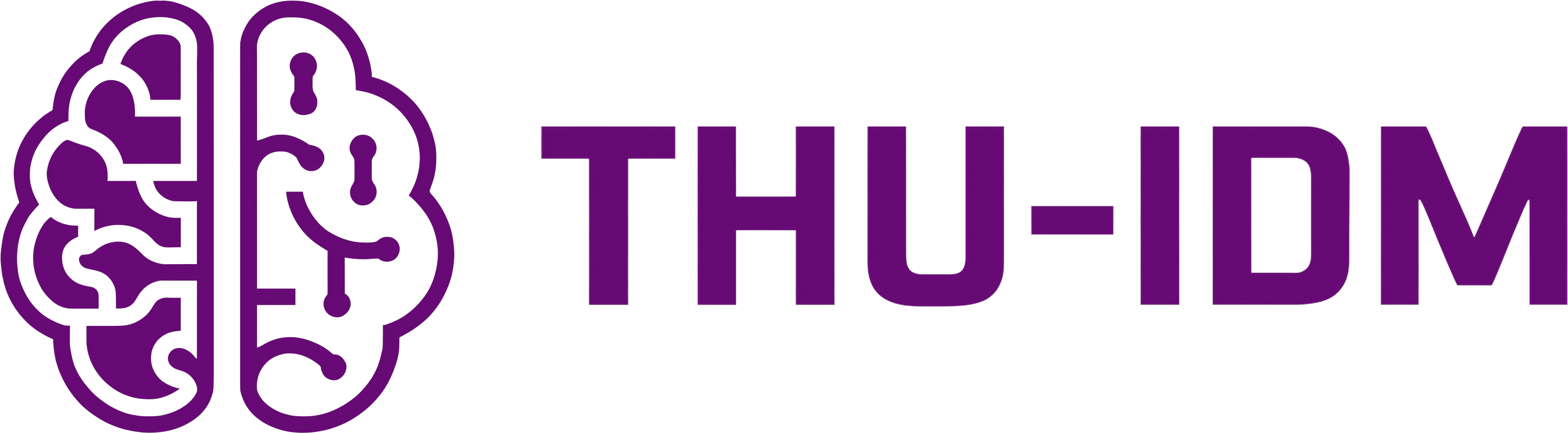}}
\sbox{\treeopsd@hylogo}{\includegraphics[height=8.6mm,trim=74.16bp 0 95.76bp 0,clip]{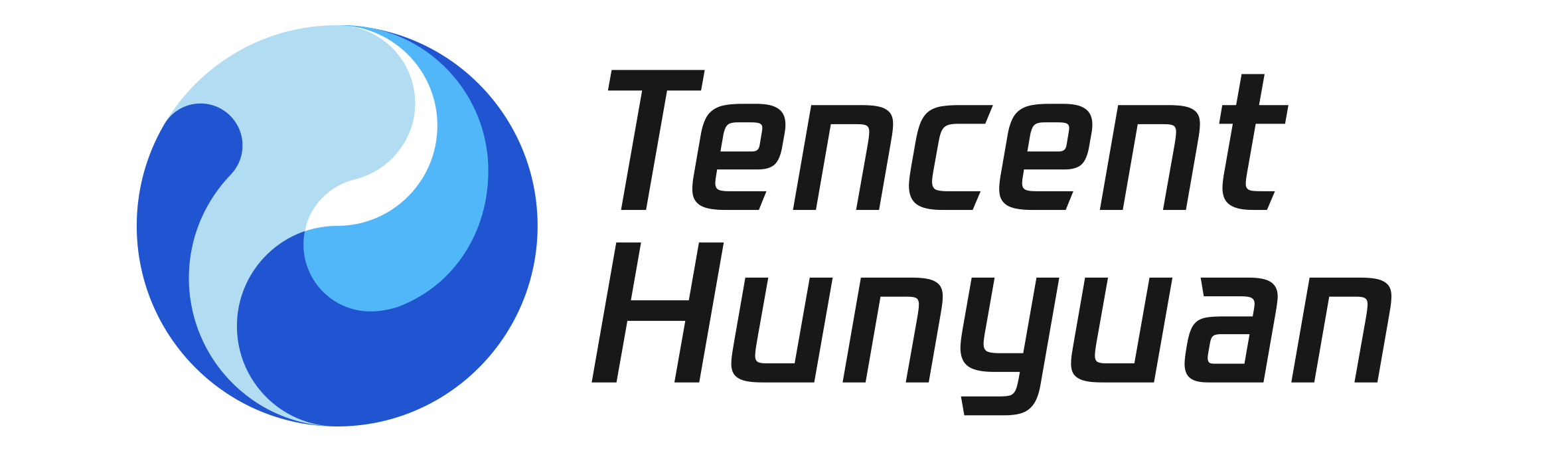}}
\newlength{\treeopsd@logogap}
\newlength{\treeopsd@logowidth}
\patchcmd{\maketitle}{\dimexpr\linewidth-42mm\relax}{\dimexpr\linewidth-\treeopsd@logowidth-2mm\relax}{}{\PackageWarning{main}{logo patch 1 failed}}
\patchcmd{\maketitle}{\dimexpr\linewidth-42mm\relax}{\dimexpr\linewidth-\treeopsd@logowidth-2mm\relax}{}{\PackageWarning{main}{logo patch 2 failed}}
\patchcmd{\maketitle}{\begin{minipage}[b]{40mm}}{\begin{minipage}[b]{\treeopsd@logowidth}}{}{\PackageWarning{main}{logo patch 3 failed}}
\patchcmd{\maketitle}{\begin{minipage}[b]{40mm}}{\begin{minipage}[b]{\treeopsd@logowidth}}{}{\PackageWarning{main}{logo patch 4 failed}}
\tcbset{thuidmcover/.append style={after skip=0pt}}
\patchcmd{\maketitle}{\endgroup\vspace{4pt}}{\endgroup}{}{\PackageWarning{main}{cover gap patch failed}}
\renewcommand{\thuidm@logobox}{%
  \raisebox{-0.5\height}{\usebox{\treeopsd@thulogo}}%
  \hspace{\treeopsd@logogap}%
  \raisebox{-0.5\height}{\usebox{\treeopsd@hylogo}}\par
  \vspace{3pt}%
  \makebox[\wd\treeopsd@thulogo][r]{\sffamily\fontsize{7.8}{9.4}\selectfont
    \href{\thuidm@weburl}{\thuidm@webtext}}\hfill\null\par}
\makeatother

\begin{document}

\begin{abstract}
Reinforcement learning with verifiable rewards has become a scalable post-training paradigm for large language models. However, training agents over long-horizon interactions presents two challenges: using only terminal rewards makes turn-level credit assignment sparse and high-variance, while sequential mini-batch updates on a fixed rollout batch introduce policy staleness as the current policy drifts from the behavior policy. Existing approaches obtain denser credit via learned critics, process reward models, or on-policy distillation, but they rely on a value network, step-level annotations, or a stronger external teacher that increase training complexity. Meanwhile, importance sampling and clipping mitigate policy drift, but they do not specify a finite target that stops sequential updates from overshooting. In this work, we introduce TreeOPSD, a critic-free, tree-aware posterior self-distillation framework for long-horizon agentic reinforcement learning. Treating each interaction turn as a tree node, TreeOPSD branches alternative continuations from shared prefixes and recursively backs up terminal rewards into implicit turn-level credit. It then maps each credit to a closed-form, bounded teacher distribution parameterized by the likelihood ratio relative to the frozen behavior policy. Minimizing the divergence between this teacher and the student policy recovers the policy-gradient direction as the tangent at the behavior policy while bounding updates toward a finite target across sequential mini-batches. Empirically, TreeOPSD improves agentic capabilities across several competitive benchmarks and model scales, especially for long-horizon tasks, e.g., raising the average accuracy of Qwen3-30B-A3B-Thinking-2507 by 4.97 points over GRPO on the Deep Research setting.
\end{abstract}

\maketitle

\section{Introduction}
\label{sec:introduction}

Reinforcement learning with verifiable rewards (RLVR) \citep{shao2024deepseekmath,deepseekai2025deepseekr1} has emerged as a central post-training paradigm for large language models (LLMs), using outcome rewards to incentivize reasoning or environment interactions on complex problem-solving tasks. While reasoning-oriented RLVR typically learns from Chain-of-Thought (CoT) rollouts, agentic RL \citep{yao2022react} extends this paradigm to long-horizon, multi-turn trajectories in which reasoning is interleaved with actions and environment observations. As the interaction horizon grows, a single terminal reward provides increasingly sparse and high-variance supervision across intermediate turns: broadcasting one trajectory-level advantage over the entire rollout over-credits unhelpful actions in successful trajectories and penalizes useful steps in failed ones \citep{arjona2019rudder,li2026salt}. At the same time, updating the policy across sequential mini-batches within each rollout batch induces policy staleness as the current policy departs from the behavior policy \citep{zheng2025stabilizing}.

To address the credit assignment problem, actor--critic methods such as Proximal Policy Optimization (PPO) \citep{schulman2017proximal} estimate token-level advantages with a learned critic, which is difficult to train accurately and susceptible to value flattening \citep{li2026valueflattening}. Process reward models (PRMs) \citep{lightman2024verify} provide step-level rewards but require process annotations, and summing them as discounted returns is vulnerable to reward hacking \citep{cheng2026stop}. On-policy distillation (OPD) \citep{agarwal2024onpolicy,gu2024minillm} provides dense token-level supervision on student rollouts at the cost of querying a stronger external teacher. Extending OPD, on-policy self-distillation (OPSD) \citep{zhao2026selfdistilled,kaur2026rethinkingopsd} constructs a self-teacher by conditioning the model on privileged context (e.g., reference solutions), which relies on task-specific heuristics and can yield unstable supervision. These limitations leave open how to derive dense intermediate credit directly from terminal verifier rewards. Sampling independent trajectories confounds all turns along each rollout, whereas multi-turn interaction exposes discrete turn-level prefixes as natural branching points. Branching alternative continuations from a shared prefix fixes the preceding history to isolate how subsequent decisions affect terminal outcomes, forming a rollout tree on which recursive backup converts terminal rewards into implicit turn-level credit without auxiliary value or reward models \citep{ji2026treesearch,zou2026trace}.

\begin{wrapfigure}{r}{0.43\textwidth}
    \centering
    \vspace{-11pt}
    \includegraphics[width=\linewidth]{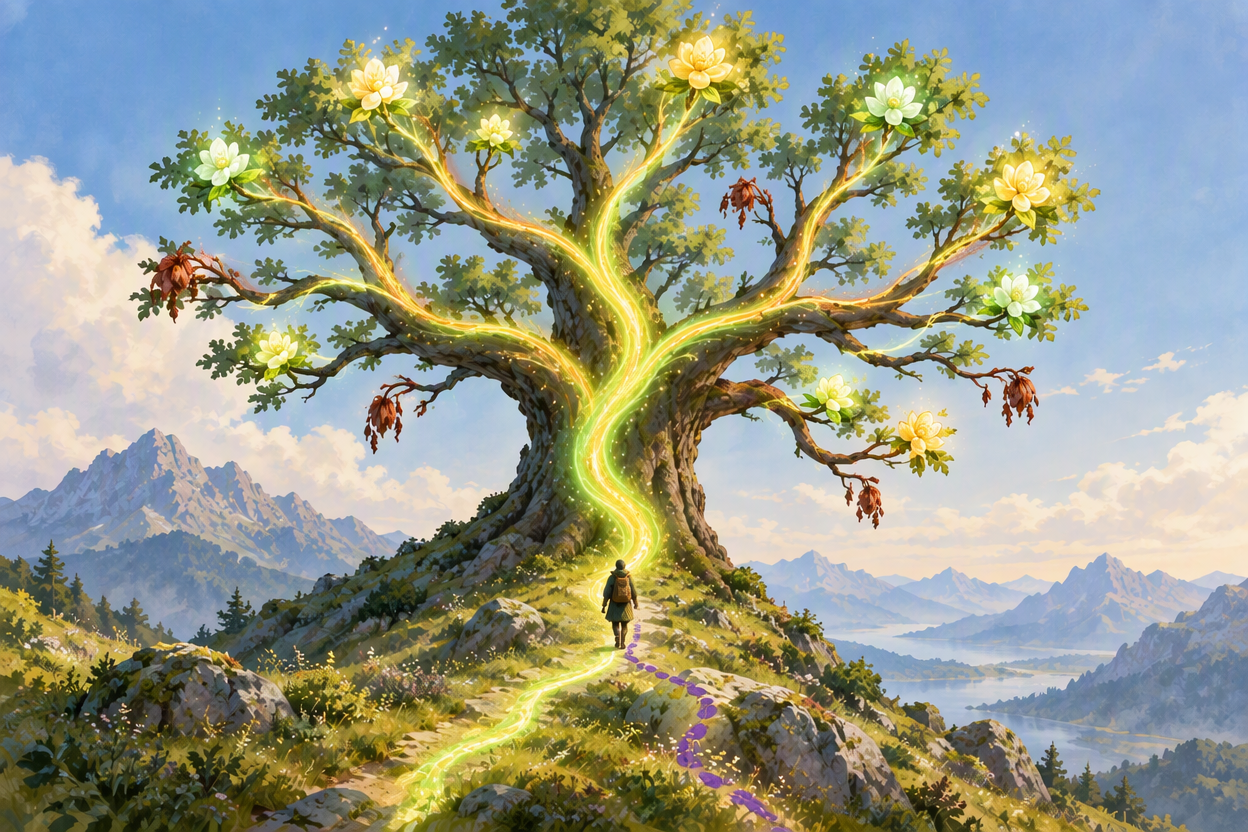}
    \vspace{-12pt}
    \caption{\textbf{Target construction:} terminal rewards at the blooming (successful) and withered (failed) branch tips are backed up through shared prefixes to define a frozen turn-level teacher along the luminous green path. \textbf{Distribution matching:} starting from the behavior-policy rollout (purple trail), the student policy progressively aligns with this outcome-derived target through posterior distillation.}
    \label{fig:treeopsd-concept}
    \vspace{-5pt}
\end{wrapfigure}

Sequential mini-batch updates on a fixed rollout batch create a separate optimization challenge. While all rollouts in the batch are sampled from a frozen behavior policy $\pi_{\theta_{\mathrm{old}}}$, each update step moves the current policy $\pi_\theta$ further away, increasing policy staleness for subsequent mini-batches. Policy-gradient methods compensate for this mismatch with importance-sampling ratios, and stabilize training through clipping \citep{schulman2017proximal} or ratio regularizers \citep{zheng2025stabilizing} as those ratios depart from one. However, such controls only constrain the importance ratio while keeping the collection-time credit fixed, so the surrogate still specifies an ascent direction rather than a finite destination \citep{qu2026listwise}. We therefore seek an objective that maps tree credit to a finite target and adapts its gradient as $\pi_\theta$ evolves, halting at the target and reversing sign after overshoot.

To this end, we introduce TreeOPSD, a critic-free posterior self-distillation framework for long-horizon agentic RL. TreeOPSD first maps backed-up turn credit into a frozen turn-level teacher distribution relative to the behavior policy, and then minimizes the divergence between this teacher and the current student policy across sequential mini-batches. Geometrically, the tangent of this distillation objective at the behavior policy recovers the exact tree policy-gradient direction, while later mini-batches scale each update by the current student--teacher gap rather than a fixed collection-time credit weight. Figure~\ref{fig:treeopsd-concept} illustrates this decoupling of target construction and distribution matching.

In summary, our main contributions are two-fold:
\begin{enumerate}[topsep=0pt,itemsep=0ex,leftmargin=3ex]
    \item We derive closed-form, bounded turn-level target distributions via an affine map of backed-up credit relative to the frozen behavior policy. Because this affine map commutes with tree backup, matching the backed-up parent target is gradient-equivalent to aggregating descendant distillation losses.
    \item We formulate a Forward-KL posterior distillation objective whose tangent at the behavior policy recovers the tree policy gradient, while its student--teacher residual vanishes at the target and reverses sign upon overshoot to stabilize sequential mini-batch optimization.
\end{enumerate}
Empirically, TreeOPSD improves agentic capabilities across competitive benchmarks and model scales, and it elicits longer-horizon agent behavior with more interaction turns in Deep Research.

\section{Preliminaries}
\label{sec:preliminaries}

\subsection{Agentic reinforcement learning}
\label{sec:agentic-rl}

We consider an LLM policy $\pi_\theta$ interacting with an environment in the ReAct setting \citep{yao2022react}. Given a prompt $x\sim\mathcal{D}$, at each turn $1\leq t\leq T$ the policy generates an $L_t$-token actor span $a_t=(a_{t,1},\ldots,a_{t,L_t})$. For nonterminal turns ($t<T$), $a_t$ comprises a thought and a tool action, after which the environment transition $P_{\mathrm{env}}$ returns an observation $o_t$ and updates the history as $\mathcal{H}_t:=(\mathcal{H}_{t-1},a_t,o_t)$ from $\mathcal{H}_0:=x$. At the terminal turn $T$, the policy emits the final answer $a_T$, yielding $\mathcal{H}_T:=(\mathcal{H}_{T-1},a_T)$. The trajectory distribution factorizes as
\begin{equation}
    \mathbb{P}_{\pi_\theta}(\mathcal{H}_T\mid x)
    =
    \underbrace{
    \pi_\theta(a_T\mid\mathcal{H}_{T-1})}_{\text{final answer}}
    \underbrace{\prod_{t=1}^{T-1}
    \pi_\theta(a_t\mid\mathcal{H}_{t-1})
    P_{\mathrm{env}}(o_t\mid\mathcal{H}_{t-1},a_t)}
    _{\text{actor--observation}},
    \label{eq:rollout-factorization}
\end{equation}
where each turn likelihood decomposes autoregressively as $\pi_\theta(a_t\mid\mathcal{H}_{t-1})=\prod_{j=1}^{L_t}\pi_\theta(a_{t,j}\mid\mathcal{H}_{t-1},a_{t,<j})$. Only policy-generated tokens contribute gradients, while observation tokens from $P_{\mathrm{env}}$ enter $\mathcal{H}_t$ as fixed context and are masked from the policy loss.

Given a reward function $R(\mathcal{H}_T)$ and a frozen reference policy $\pi_{\mathrm{ref}}$, the standard KL-regularized agentic RL objective \citep{ouyang2022training,shao2024deepseekmath} with reference weight $\beta_{\mathrm{KL}}\geq 0$ is
\begin{equation}
    \max_\theta\ 
    \mathbb{E}_{\substack{x\sim\mathcal{D}\\
    \mathcal{H}_T\sim\mathbb{P}_{\pi_\theta}(\cdot\mid x)}}
    [R(\mathcal{H}_T)]
    -
    \beta_{\mathrm{KL}}\,
    \mathbb{E}_{x\sim\mathcal{D}}
    \left[
        D_{\mathrm{KL}}
        \left(
            \mathbb{P}_{\pi_\theta}(\cdot\mid x)\Vert \mathbb{P}_{\pi_{\mathrm{ref}}}(\cdot\mid x)
        \right)
    \right].
    \label{eq:agentic-objective}
\end{equation}
Following recent advances \citep{yu2026dapo,zou2026trace,qu2026listwise}, we focus on rule-based outcome rewards that are binary or sparse ($R(\mathcal{H}_T)\in[0,1]$) without an explicit reference penalty ($\beta_{\mathrm{KL}}=0$), where intermediate turns $t<T$ receive no direct supervision.

\subsection{Initialize-then-expand rollout trees}
\label{sec:rollout-trees}

To estimate intermediate turn credit from terminal rewards alone, tree-based policy optimization branches alternative continuations from shared prefixes via an initialize-then-expand sampling scheme \citep{hou2025treerl,ji2026treesearch,dong2026arpo,zong2026at2po,zou2026trace}. For each prompt $x$, the frozen behavior policy $\pi_{\theta_{\mathrm{old}}}$ first samples $M$ complete trajectories $\mathcal{H}_{m,T_m}$, where $T_m$ is the terminal-turn index of trajectory $m$. From each trajectory $m$, the sampler uniformly draws a multiset $\mathcal{A}_m=\{t_{m,1},\ldots,t_{m,N}\}$ of $N$ nonterminal anchor indices $1\leq t_{m,n}<T_m$ (or $t_{m,n}=0$ if $T_m=1$), and samples one fresh terminal suffix $S_{m,n}$ from each prefix $\mathcal{H}_{m,t_{m,n}}$:
\begin{equation}
    \begin{aligned}
        \mathcal{H}_{m,T_m}
        &\sim \mathbb{P}_{\pi_{\theta_{\mathrm{old}}}}(\cdot\mid x),
        &&m=1,\ldots,M,\\
        S_{m,n}
        &\sim \mathbb{P}_{\pi_{\theta_{\mathrm{old}}}}
        (\cdot\mid\mathcal{H}_{m,t_{m,n}}),
        &&t_{m,n}\in\mathcal{A}_m,\quad n=1,\ldots,N.
    \end{aligned}
    \label{eq:tree-sampling}
\end{equation}
Retaining the $M$ initial rollouts alongside their $MN$ branched continuations yields $K:=M(N+1)$ terminal trajectories. Because a uniformly branched continuation starts mid-trajectory, its expected generation cost in tokens or tool calls is roughly half that of a full rollout \citep{ji2026treesearch,zou2026trace}, so an $(M,N)$ tree matches the expected rollout budget of $G$ full trajectories in Group Relative Policy Optimization (GRPO) \citep{shao2024deepseekmath} when $M(1+N/2)\approx G$. Merging shared prefixes across the $K$ trajectories produces a rollout tree with node set $\mathcal{T}_x$, where shared turns appear once while independently sampled continuations form distinct nodes. Each node $i\in\mathcal{T}_x$ represents an $L_i$-token actor span $a_i$ conditioned on pre-turn history $h_i$, with token-level context $h_{i,j}:=(h_i,a_{i,<j})$ for $1\leq j\leq L_i$ and sampled child set $\mathcal{C}_i$. Each terminal trajectory $k\in\{1,\ldots,K\}$ ends at a leaf node $i_k\in\mathcal{T}_x$ and receives verifier reward $R_k$.

\subsection{Tree-based turn credit}
\label{sec:tree-credit}

Following common practice in tree-based policy optimization \citep{hou2025treerl,ji2026treesearch,dong2026arpo,zong2026at2po,zou2026trace}, terminal outcomes on $\mathcal{T}_x$ are propagated upward to assign credit to each turn. First, the $K$ terminal rewards for prompt $x$ are standardized into trajectory advantages $A_k$, as in group-based RLVR:
\begin{equation}
    A_k
    :=
    \frac{
        R_k-\operatorname{mean}(\{R_{k'}\}_{k'=1}^{K})
    }{
        \operatorname{std}(\{R_{k'}\}_{k'=1}^{K})
    },
    \qquad k=1,\ldots,K,
    \label{eq:standardized-reward}
\end{equation}
using the sample standard deviation (with $A_k:=0$ when all rewards are equal). These leaf advantages are then recursively backed up through $\mathcal{T}_x$ using behavior-policy likelihood weights:
\begin{equation}
    \begin{aligned}
    \omega_{i,c}
    &:=
    \frac{\pi_{\theta_{\mathrm{old}}}(a_c\mid h_c)}
         {\sum_{c'\in\mathcal{C}_i}\pi_{\theta_{\mathrm{old}}}(a_{c'}\mid h_{c'})},
    &
    V_i
    &:=
    \begin{cases}
        A_k, & i=i_k,\\
        \sum_{c\in\mathcal{C}_i}\omega_{i,c}V_c,
        & \text{otherwise},
    \end{cases}
    \end{aligned}
    \label{eq:policy-backup}
\end{equation}
where $\pi_{\theta_{\mathrm{old}}}(a_c\mid h_c)$ is the behavior-policy likelihood of child span $a_c$ given the shared post-observation history $h_c=(h_i,a_i,o_i)$ of parent turn $i$. Because $V_i$ already aggregates the outcomes of all branches below turn $i$, each turn $i\in\mathcal{T}_x$ enters the gradient once rather than being repeated across the $K$ trajectories. With $L_x:=\sum_{i\in\mathcal{T}_x}L_i$ denoting the total number of unique actor tokens in $\mathcal{T}_x$, the prompt-level tree policy-gradient direction is
\begin{equation}
    g_x^{\mathrm{tree}}
    :=
    \frac{1}{L_x}
    \sum_{i\in\mathcal{T}_x}
    \sum_{j=1}^{L_i}
    V_i
    \left.
    \nabla_\theta
    \log\pi_\theta(a_{i,j}\mid h_{i,j})
    \right|_{\theta=\theta_{\mathrm{old}}}.
    \label{eq:tree-pg-direction}
\end{equation}
Eq.~\ref{eq:tree-pg-direction} assigns the weight $V_i/L_x$ to every policy-generated token in turn $i$, providing the exact tree policy-gradient ascent direction at the behavior policy $\pi_{\theta_{\mathrm{old}}}$.

\section{TreeOPSD}
\label{sec:rl-to-tree-projection}

While the tree policy gradient $g_x^{\mathrm{tree}}$ in Eq.~\ref{eq:tree-pg-direction} is exact at the on-policy point $\pi_\theta=\pi_{\theta_{\mathrm{old}}}$, practical RLVR training partitions each rollout batch into sequential mini-batches. Once earlier mini-batches update $\theta$, the current policy $\pi_\theta$ drifts away from the behavior policy $\pi_{\theta_{\mathrm{old}}}$ that generated the tree. Continuing to apply the collection-time credit $V_i$ as a fixed linear weight ignores this off-policy drift and specifies only an initial ascent direction rather than a finite destination \citep{qu2026listwise}. Conversely, multiplying token gradients by raw importance ratios $\pi_\theta/\pi_{\theta_{\mathrm{old}}}$ amplifies variance on long turns \citep{ionides2008truncated,liu2018breaking,liu2020understanding}, while ratio clipping truncates the update without defining a target distribution where optimization should converge \citep{schulman2017proximal,zheng2025stabilizing,qu2026listwise}.

\begin{figure}[t]
    \centering
    \includegraphics[width=\linewidth]{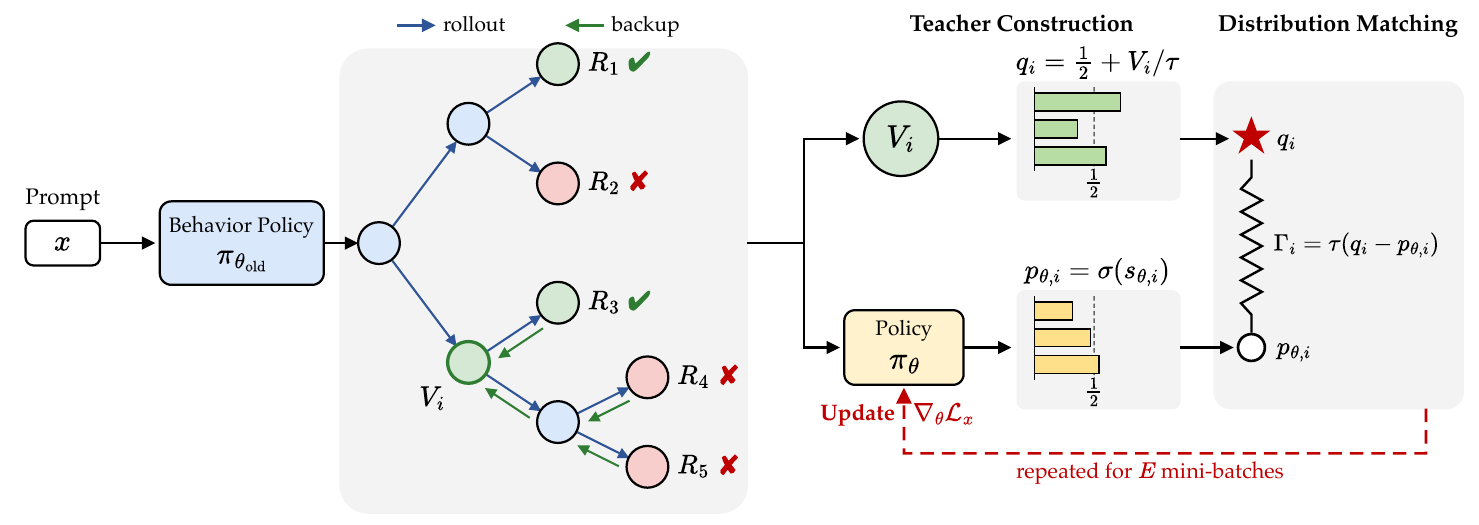}
    \vspace{-10pt}
    \caption{\textbf{TreeOPSD framework.} The behavior policy $\pi_{\theta_{\mathrm{old}}}$ generates a shared-prefix rollout tree, and terminal verifier rewards $R_k$ are backed up into turn credits $V_i$. Each credit defines a frozen Bernoulli teacher $q_i=\tfrac12+V_i/\tau$, and each bar shows one turn against the behavior level $\tfrac12$ (dashed). In each of the $E$ sequential mini-batches, the student posterior $p_{\theta,i}=\sigma(s_{\theta,i})$ is recomputed under the current policy $\pi_\theta$, and the spring-like ascent weight $\Gamma_i=\tau(q_i-p_{\theta,i})$ pulls it toward $q_i$, vanishing at the teacher and reversing sign after overshoot.}
    \label{fig:treeopsd-overview}
\end{figure}

These limitations arise because linear policy-gradient surrogates entangle credit assignment with policy updating instead of specifying an explicit target distribution. Drawing on Policy Mirror Descent (PMD) and variational policy optimization \citep{beck2003mirror,abdolmaleki2018maximum,vaswani2022general,qu2026listwise}, we decouple each iteration into two steps: first solve a KL-regularized proximal improvement problem around $\pi_{\theta_{\mathrm{old}}}$ to convert backed-up turn credits into a frozen target distribution, and then project the evolving policy $\pi_\theta$ onto that target across sequential mini-batches. This decoupling addresses both bottlenecks: target construction maps backed-up credits into tree-consistent, bounded targets at every depth, while distribution matching replaces high-variance importance ratios with a convex projection whose gradient vanishes once the student reaches the target. Writing $Q$ for a candidate target distribution, $\mathcal{J}(Q)$ for a proximal objective anchored at $\pi_{\theta_{\mathrm{old}}}$, $P_\theta$ for the current student distribution, and $D(Q\Vert P_\theta)$ for a statistical divergence, the two steps are
\begin{equation}
    \underbrace{
    Q^\star
    =
    \arg\max_{Q}
    \mathcal{J}(Q)
    }_{\text{target construction}}
    \qquad
    \underbrace{
    \theta'
    =
    \arg\min_\theta
    D\!\left(Q^\star\Vert P_{\theta}\right)
    }_{\text{distribution matching}}.
    \label{eq:target-matching-principle}
\end{equation}
Target construction is executed once per rollout tree from $\pi_{\theta_{\mathrm{old}}}$, holding $Q^\star$ fixed across the batch, whereas distribution matching re-evaluates $D(Q^\star\Vert P_\theta)$ at every mini-batch step as $\pi_\theta$ evolves (Figure~\ref{fig:treeopsd-overview}).

\subsection{Behavior-relative turn distributions}
\label{sec:target-and-projection}

As established in Section~\ref{sec:tree-credit}, branching from shared prefixes naturally converts sparse terminal outcomes into dense turn-level credit $V_i$ without an external teacher or learned critic. Unlike token-level distillation where an external teacher supplies full-vocabulary distributions at every token, here the supervision signal is a single turn-level scalar $V_i$ attached to the entire realized turn span $a_i$ of length $L_i$. To match this turn-level credit granularity, $P_\theta$ at node $i$ must represent how strongly the complete turn $a_i$ is reinforced or suppressed relative to the behavior policy $\pi_{\theta_{\mathrm{old}}}$ that sampled it, using only the logged probabilities of the sampled tokens rather than vocabulary-wide behavior logits \citep{qi2026rethinking}.

Directly multiplying token likelihood ratios across a multi-token turn compounds off-policy discrepancies exponentially with $L_i$ \citep{liu2018breaking,liu2020understanding}, causing severe variance and length bias between short tool calls and long reasoning turns. We therefore parameterize the turn-level policy shift by the per-token average log-likelihood ratio \citep{meng2024simpo}, or equivalently the geometric-mean likelihood ratio \citep{zheng2025gspo}, which normalizes this shift by turn length:
\begin{equation}
    s_{\theta,i}
    :=
    \frac1{L_i}\sum_{j=1}^{L_i}
    \log
    \frac{\pi_\theta(a_{i,j}\mid h_{i,j})}
         {\pi_{\theta_{\mathrm{old}}}(a_{i,j}\mid h_{i,j})},
    \qquad
    r_{\theta,i}:=e^{s_{\theta,i}}
    =
    \left(
    \frac{\pi_\theta(a_i\mid h_i)}
         {\pi_{\theta_{\mathrm{old}}}(a_i\mid h_i)}
    \right)^{1/L_i}.
    \label{eq:turn-score}
\end{equation}
At the on-policy point $\pi_\theta=\pi_{\theta_{\mathrm{old}}}$, the log-ratio score is zero and $r_{\theta,i}=1$. As $\pi_\theta$ drifts during sequential updates, $r_{\theta,i}$ rises above one when turn $i$ becomes more likely than under $\pi_{\theta_{\mathrm{old}}}$ and falls below one when it becomes less likely. Because normalizing $r_{\theta,i}$ over the full continuation space requires an intractable partition function, we follow noise-contrastive and relative density-ratio estimation \citep{yamada2013relative,goodfellow2014generative,cranmer2015approximating,kobayashi2021pporpe,chen2024noise} and form a two-point conditional distribution between $\pi_\theta$ and the behavior baseline $\pi_{\theta_{\mathrm{old}}}$ (whose ratio to itself is $1$). Normalizing the pair $(r_{\theta,i}, 1)$ with $\sigma(s):=(1+e^{-s})^{-1}$ yields the binary posterior
\begin{equation}
    p_{\theta,i}
    :=
    \frac{r_{\theta,i}}{1+r_{\theta,i}}
    =
    \sigma(s_{\theta,i}),
    \qquad
    P_{\theta,i}:=\operatorname{Ber}(p_{\theta,i}).
    \label{eq:turn-distribution}
\end{equation}
Normalizing in likelihood-ratio space makes $P_{\theta,i}$ invariant to the raw sequence probability of $a_i$ and eliminates any partition function over unsampled continuations. This odds map bijectively converts $s_{\theta,i}$ into a Bernoulli distribution whose log-odds equal $s_{\theta,i}$, mapping the on-policy anchor $r_{\theta,i}=1$ to $p_{\theta,i}=1/2$ and reflecting $p_{\theta,i}$ symmetrically around $1/2$ when $\pi_\theta$ and $\pi_{\theta_{\mathrm{old}}}$ are swapped.

\subsection{Teacher construction}
\label{sec:soft-values}

Given the student posterior $P_{\theta,i}=\operatorname{Ber}(p_{\theta,i})$, we construct a turn-level teacher $Q_i^\star=\operatorname{Ber}(q_i)$ in the same Bernoulli family. Notice from Eqs.~\ref{eq:tree-pg-direction} and~\ref{eq:turn-score} that the on-policy tree gradient at node $i$ is the gradient of the linear surrogate $V_i s_{\theta,i}$ (up to the weight $L_i/L_x$), while inverting the sigmoid map $p=\sigma(s)$ in Eq.~\ref{eq:turn-distribution} expresses the log-ratio score of any candidate Bernoulli distribution $\operatorname{Ber}(q)$ as its log-odds $s(q)=\sigma^{-1}(q)=\log\frac{q}{1-q}$. Thus, in Bernoulli coordinates, the linear policy-improvement term $V_i s$ takes the exact form $V_i\log\frac{q}{1-q}$. Following PMD and KL-regularized policy optimization \citep{ouyang2022training,abdolmaleki2018maximum,vaswani2022general}, we regularize this linear improvement with a KL trust-region penalty from the on-policy behavior distribution $\operatorname{Ber}(1/2)$ (where $s=0$) at temperature $\tau>0$:
\begin{equation}
    \mathcal{J}_i\!\left(\operatorname{Ber}(q)\right)
    :=
        V_i\log\frac{q}{1-q}
        -
        \tau D_{\mathrm{KL}}\!\left(
            \operatorname{Ber}(\tfrac12)\Vert\operatorname{Ber}(q)
        \right).
    \label{eq:bernoulli-target}
\end{equation}
Choosing the forward-direction KL penalty from the behavior anchor $\operatorname{Ber}(1/2)$ to the candidate teacher $\operatorname{Ber}(q)$ is key to preserving tree credit across depths. In log-odds coordinates, this KL penalty provides a smooth local quadratic trust region around zero whose derivative is linear in $q$ and globally bounded in $(-1/2,1/2)$, so the concavity of $\mathcal{J}_i$ yields a closed-form teacher probability $q_i$ that is affine in $V_i$, together with the target likelihood ratio $r_i^\star$ (Proposition~B.2 in Appendix~\ref{app:credit-details}):
\begin{equation}
    q_i
    :=
    \frac12+\frac{V_i}{\tau},
    \qquad
    r_i^\star
    :=
    \frac{q_i}{1-q_i}
    =
    \frac{\tau+2V_i}{\tau-2V_i}.
    \label{eq:target-ratio}
\end{equation}
Positive credit places the target above the behavior policy ($q_i>1/2$, $r_i^\star>1$), negative credit places it below ($q_i<1/2$, $r_i^\star<1$), and zero credit leaves the behavior policy unchanged ($q_i=1/2$, $r_i^\star=1$).

To ensure that every turn target remains strictly valid and bounded without ad-hoc clipping, we calibrate $\tau$ from the group size $K$. By Samuelson's inequality \citep{samuelson1968deviant}, the standardized terminal advantages in Eq.~\ref{eq:standardized-reward} are bounded by $(K-1)/\sqrt{K}$, and the convex tree backup in Eq.~\ref{eq:policy-backup} preserves this bound at every internal node. Setting $\tau=(K-1)/(\rho\sqrt{K})$ for a target radius $\rho\in(0,1/2)$ confines every $q_i$ to $[1/2-\rho,1/2+\rho]$ and every target ratio to the depth-independent interval
\begin{equation}
    r_i^\star\in\left[\frac{1-2\rho}{1+2\rho},\,\frac{1+2\rho}{1-2\rho}\right].
    \label{eq:teacher-trust-region}
\end{equation}
Moreover, as proved in Appendix~\ref{app:credit-details} and validated in Section~\ref{sec:target-geometry}, the affine teacher in Eq.~\ref{eq:target-ratio} is the unique symmetric radius-calibrated map that commutes with convex tree backup.

\subsection{Distribution matching and off-policy analysis}
\label{sec:distribution-matching}

With the teacher $Q_i^\star=\operatorname{Ber}(q_i)$ frozen at rollout, the second step in Eq.~\ref{eq:target-matching-principle} projects the evolving student $P_{\theta,i}=\operatorname{Ber}(p_{\theta,i})$ onto $Q_i^\star$ across sequential mini-batches. A principled choice of projection divergence $D$ follows directly from the proximal objective in Eq.~\ref{eq:bernoulli-target}: by Bregman and Fenchel--Young duality \citep{banerjee2005clustering,blondel2020fenchel}, the optimality gap between $Q_i^\star$ and any student $P_{\theta,i}$ is exactly the Forward KL divergence $\mathcal{J}_i(Q_i^\star)-\mathcal{J}_i(P_{\theta,i})=\tau D_{\mathrm{KL}}(Q_i^\star\Vert P_{\theta,i})$ \citep{agarwal2024onpolicy,qu2026listwise}. Weighting each turn's Forward KL by $L_i/L_x$ to match the token normalization in Eq.~\ref{eq:tree-pg-direction} yields the TreeOPSD loss:
\begin{equation}
    \mathcal{L}_{x}(\theta)
    :=
    \frac{\tau}{L_x}\sum_{i\in\mathcal{T}_x}L_iD_{\mathrm{KL}}\!\left(Q_i^\star\Vert P_{\theta,i}\right)
    =
    \frac{1}{L_x}\sum_{i\in\mathcal{T}_x}L_i
    \left[
        -V_i s_{\theta,i}
        +\tau D_{\mathrm{KL}}\!\left(\operatorname{Ber}(\tfrac12)\Vert P_{\theta,i}\right)
    \right]
    +C_x,
    \label{eq:treeopsd-loss}
\end{equation}
where $C_x$ is independent of $\theta$. The second expression in Eq.~\ref{eq:treeopsd-loss} shows why Forward-KL matching stabilizes off-policy updates: every step that reduces the distillation loss $\mathcal{L}_x(\theta)$ monotonically increases the tree-weighted proximal RL objective (Appendix~\ref{app:proof-tangent}), and the loss combines linear credit ascent $-V_i s_{\theta,i}$ with a convex restoring penalty around $\pi_{\theta_{\mathrm{old}}}$ that activates only as $\pi_\theta$ drifts away from the behavior policy.

We now establish two structural theorems showing how Eq.~\ref{eq:treeopsd-loss} preserves exact turn-level credit across the tree (Theorem~\ref{thm:exact-marginalization}) and corrects off-policy drift across sequential mini-batches (Theorem~\ref{thm:pg-tangent}).

\begin{figure}[t]
    \centering
    \includegraphics[width=\linewidth]{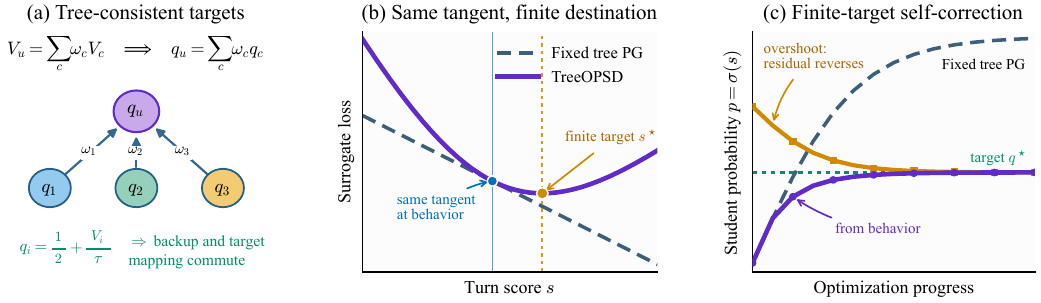}
    \vspace{-10pt}
    \caption{\textbf{From backed-up credit to a finite, self-correcting target.} (a) The affine credit-to-target map commutes with likelihood-weighted tree backup, so the teacher $q_u$ at a shared turn equals the weighted average of its child teachers. (b) In log-ratio score space $s_{\theta,i}$, TreeOPSD and fixed tree policy gradient share the same tangent at the behavior policy ($s=0$), but only TreeOPSD has a finite minimizer $s^\star=\log r^\star$. (c) The student probability $p_{\theta,i}=\sigma(s_{\theta,i})$ converges toward the target $q^\star$ from either side because the residual reverses sign after overshoot, whereas fixed credit keeps pushing in the initial direction.}
    \label{fig:conceptual-geometry}
\end{figure}

\begin{theorem}[Tree-consistent targets and descendant loss aggregation]
\label{thm:exact-marginalization}
Let $\mathbb{E}_{k\mid i}$ denote the expectation over terminal leaves $i_k$ below turn $i$ under the recursive backup weights in Eq.~\ref{eq:policy-backup}. For any student probability $p_{\theta,i}\in(0,1)$, the leaf-level Forward KL losses decompose as
\begin{equation}
    \mathbb{E}_{k\mid i}
    \left[
        D_{\mathrm{KL}}\!\left(
            \operatorname{Ber}(q_{i_k})\Vert\operatorname{Ber}(p_{\theta,i})
        \right)
    \right]
    =
    \mathbb{E}_{k\mid i}
    \left[
        D_{\mathrm{KL}}\!\left(
            \operatorname{Ber}(q_{i_k})\Vert\operatorname{Ber}(q_i)
        \right)
    \right]
    +
    D_{\mathrm{KL}}\!\left(
        \operatorname{Ber}(q_i)\Vert\operatorname{Ber}(p_{\theta,i})
    \right).
    \label{eq:loss-decomposition}
\end{equation}
Because the first term on the right-hand side is independent of $\theta$, matching the single backed-up parent teacher $Q_i^\star=\operatorname{Ber}(q_i)$ yields the exact same gradient on turn $i$ as aggregating the distillation losses of all descendant leaves below $i$. Equivalently, the affine map $q_i=1/2+V_i/\tau$ commutes with tree backup and coincides with the Forward-KL barycenter \citep{banerjee2005clustering} of its child teachers:
\begin{equation}
    q_i
    =
    \frac12+\frac{V_i}{\tau}
    =
    \sum_{c\in\mathcal{C}_i}\omega_{i,c}q_c
    =
    \arg\min_{q\in(0,1)}
    \sum_{c\in\mathcal{C}_i}\omega_{i,c}
    D_{\mathrm{KL}}\!\left(
        \operatorname{Ber}(q_c)\Vert\operatorname{Ber}(q)
    \right).
    \label{eq:tree-consistent-teacher}
\end{equation}
\end{theorem}
Theorem~\ref{thm:exact-marginalization} guarantees lossless credit compression on the rollout tree: the parent target $q_i$ is identical whether descendant credits are mapped before backup or the backed-up credit $V_i$ is mapped afterward (Figure~\ref{fig:conceptual-geometry}(a)), allowing Eq.~\ref{eq:treeopsd-loss} to score each unique turn once without repeating shared prefixes across leaves. Proofs are given in Appendix~\ref{app:credit-details}--\ref{app:proof-barycenter}.

\begin{theorem}[On-policy tangent and off-policy drift correction]
\label{thm:pg-tangent}
For every token $j$ in turn $i\in\mathcal{T}_x$, differentiating Eq.~\ref{eq:treeopsd-loss} with respect to $\log\pi_\theta(a_{i,j}\mid h_{i,j})$ yields
\begin{equation}
    \frac{\partial\mathcal{L}_{x}}
    {\partial\log\pi_\theta(a_{i,j}\mid h_{i,j})}
    =
    \frac{\tau}{L_x}(p_{\theta,i}-q_i)
    =
    -\frac{1}{L_x}
    \left[
        V_i-\tau\left(p_{\theta,i}-\frac12\right)
    \right].
    \label{eq:projection-residual}
\end{equation}
Here the bracketed ascent weight $\Gamma_i(s_{\theta,i})$ subtracts an off-policy drift correction from the on-policy credit $V_i$. Expressed in terms of the current and target likelihood ratios $r_{\theta,i}$ and $r_i^\star$, it takes the self-normalized form
\begin{equation}
    \Gamma_i(s_{\theta,i})
    :=
    V_i-\tau\left(p_{\theta,i}-\frac12\right)
    =
    \tau
    \frac{r_i^\star-r_{\theta,i}}
         {(1+r_i^\star)(1+r_{\theta,i})}.
    \label{eq:effective-credit}
\end{equation}
Across sequential mini-batch updates, the ascent gradient $-\nabla_\theta\mathcal{L}_x(\theta)$ satisfies three properties:
\begin{enumerate}[label=\upshape(\roman*),widest=iii,leftmargin=*,topsep=2pt,itemsep=1pt]
    \item \textbf{Exact on-policy tangent:} At the behavior policy $\pi_\theta=\pi_{\theta_{\mathrm{old}}}$, we have $r_{\theta,i}=1$ and $p_{\theta,i}=1/2$, so the drift correction vanishes and $-\nabla_\theta\mathcal{L}_x(\theta_{\mathrm{old}})=g_x^{\mathrm{tree}}$ exactly recovers Eq.~\ref{eq:tree-pg-direction}.
    \item \textbf{Bounded additive drift cancellation:} As $\pi_\theta$ drifts away from $\pi_{\theta_{\mathrm{old}}}$ on subsequent mini-batches, the additive correction $-\tau(p_{\theta,i}-1/2)$ counteracts the policy shift, remaining Lipschitz in $s_{\theta,i}$ and uniformly bounded for all $r_{\theta,i}\in(0,\infty)$ without the exponential variance of multiplicative importance ratios \citep{ionides2008truncated,liu2018breaking}.
    \item \textbf{Finite-target equilibrium and sign reversal:} $\Gamma_i(s_{\theta,i})$ vanishes once $r_{\theta,i}$ reaches $r_i^\star$ and reverses sign if earlier mini-batches push the turn past $r_i^\star$, whereas fixed-credit updates keep $V_i$ unchanged regardless of drift.
\end{enumerate}
\end{theorem}
Figures~\ref{fig:conceptual-geometry}(b)--(c) illustrate these dynamics, and Appendix~\ref{app:proof-tangent} proves Theorem~\ref{thm:pg-tangent} along with scalar contraction and parameter-level drift bounds between TreeOPSD and fixed-credit updates.

\subsection{Practical optimization}
\label{sec:sequential-optimization}

Algorithm~\ref{alg:treeopsd} implements Eq.~\ref{eq:target-matching-principle} in two stages per training iteration. We first freeze the behavior policy $\pi_{\theta_{\mathrm{old}}}$, sample a batch $\mathcal{B}$ of $B$ prompts, and partition it into $E$ disjoint mini-batches $\mathcal{B}=\bigcup_{e=1}^{E}\mathcal{B}_e$, corresponding to $E$ sequential optimizer blocks. In the target-construction stage, we generate the $(M,N)$ rollout tree $\mathcal{T}_x$, compute backed-up credits $V_i$, and set the frozen teachers $q_i=1/2+V_i/\tau$ for all $x\in\mathcal{B}$. In the distribution-matching stage, each block $e=1,\ldots,E$ evaluates its mini-batch $\mathcal{B}_e$ under the current policy $\pi_\theta$, recomputes the off-policy scores $s_{\theta,i}$ and posteriors $p_{\theta,i}$, and backpropagates the token loss-gradient weights $\tau(p_{\theta,i}-q_i)/L_x=-\Gamma_i(s_{\theta,i})/L_x$ to update $\theta$.

\begin{algorithm}[t]
  \caption{TreeOPSD training}
  \label{alg:treeopsd}
  \KwInput{
    Prompt dataset $\mathcal{D}$;
    initial policy $\pi_\theta$;
    prompt batch size $B$;
    initial-rollout count $M$;
    branch count $N$;
    optimizer-block count $E$;
    target radius $\rho$.
  }
  \KwOutput{Updated policy $\pi_\theta$.}
  \ForEach{training iteration}{
    Freeze $\pi_{\theta_{\mathrm{old}}}\leftarrow\operatorname{stopgrad}(\pi_\theta)$\;
    Sample $\mathcal{B}\subset\mathcal{D}$ with $|\mathcal{B}|=B$ and partition into disjoint mini-batches $\mathcal{B}=\bigcup_{e=1}^{E}\mathcal{B}_e$\;
    Set $K=M(N+1)$ and $\tau=(K-1)/(\rho\sqrt K)$\;
    \tcp{\textcolor{uclablue}{Construct rollout trees and turn teachers}}
    \ForEach{prompt $x\in\mathcal{B}$}{
      Sample and retain $M$ complete rollouts $\mathcal{H}_{m,T_m}$ from $\pi_{\theta_{\mathrm{old}}}(\cdot\mid x)$\;
      \ForEach{retained rollout $m=1,\ldots,M$}{
        Draw a multiset $\mathcal{A}_m=\{t_{m,1},\ldots,t_{m,N}\}$ of $N$ branch anchors from rollout $m$\;
        \ForEach{$t_{m,n}\in\mathcal{A}_m$}{
          Sample one fresh terminal suffix $S_{m,n}$ from prefix $\mathcal{H}_{m,t_{m,n}}$\;
        }
      }
      Merge shared prefixes into $\mathcal{T}_x$ and score its $K$ terminal trajectories\;
      Standardize terminal rewards once to $A_k$ and back up all $V_i$ by Eq.~\ref{eq:policy-backup}\;
      Set $q_i=1/2+V_i/\tau$ for every $i\in\mathcal{T}_x$\;
    }
    \tcp{\textcolor{uclablue}{Match the evolving student policy to the frozen teachers}}
    \For{$e=1$ \KwTo $E$}{
      \ForEach{prompt $x\in\mathcal{B}_e$}{
        \ForEach{unique turn $i\in\mathcal{T}_x$}{
          Recompute $s_{\theta,i}$ and $p_{\theta,i}=\sigma(s_{\theta,i})$ under the current policy $\pi_\theta$\;
          Assign loss-gradient weight $\tau(p_{\theta,i}-q_i)/L_x$ to every actor token in turn $i$\;
        }
      }
      Backpropagate the prompt-averaged token gradients and take one optimizer step on $\theta$\;
    }
  }
\end{algorithm}

\section{Experiments}
\label{sec:experiments}

\subsection{Experimental setup}
\label{sec:experimental-setup}

\paragraph{Datasets and models.} We evaluate TreeOPSD on two representative multi-turn settings. (1)~\textbf{Multi-Hop QA:} trains on the HotpotQA (Hotpot) training split \citep{yang2018hotpotqa} and evaluates on its validation split, 2WikiMultiHopQA (2Wiki) \citep{ho2020constructing}, MuSiQue \citep{trivedi2022musique}, and Bamboogle \citep{press2023measuring} with a local E5 retrieval server \citep{wang2022text} built over a Wikipedia dump \citep{karpukhin2020dense}. (2)~\textbf{Deep Research:} trains on the 6,102 unique questions released with OpenResearcher \citep{li2026openresearcher} and evaluates on BrowseComp-Plus (BC+) \citep{chen2025browsecompplus} with closed-corpus retrieval, then tests open-web transfer on BrowseComp (BC) \citep{wei2025browsecomp}, GAIA \citep{mialon2024gaia}, and xbench-DeepSearch (xbench) \citep{chen2025xbench}. Main experiments use Qwen3-4B-Thinking-2507 and Qwen3-30B-A3B-Thinking-2507 \citep{yang2025qwen3}.

\paragraph{Baselines.} We compare TreeOPSD with four baselines. (1)~\textbf{ReAct} \citep{yao2022react}: evaluates the base model in the same multi-turn agent and tool scaffold. (2)~\textbf{GRPO} \citep{shao2024deepseekmath}: normalizes terminal rewards over complete rollouts from each prompt. (3)~\textbf{GSPO} \citep{zheng2025gspo}: replaces token-level importance ratios with a sequence-level ratio while retaining group-relative terminal advantages. (4)~\textbf{TreePO}: uses the same initialize-then-expand trees, likelihood-weighted turn credits $V_i$, prompt normalization, and optimizer blocks as TreeOPSD, but applies the fixed ascent coefficient $V_i/L_x$ throughout optimization instead of matching a finite posterior target. GRPO and GSPO use $G=8$ rollouts per prompt, and both tree methods use $M=4$ initial rollouts and $N=2$ continuations per initial rollout to match that budget under expectation.

\paragraph{Metrics.} We report exact match (EM) in both settings. ``Avg'' is the unweighted mean over HotpotQA, 2WikiMultiHopQA, MuSiQue, and Bamboogle for Multi-Hop QA, and over BrowseComp-Plus, BrowseComp, GAIA, and xbench-DeepSearch for Deep Research. We evaluate every five updates through step 80 on Multi-Hop QA and step 70 on BrowseComp-Plus. Appendix~\ref{app:experimental-details} gives the data splits, rewards, training settings, and evaluation protocol.

\subsection{Main results}
\label{sec:main-comparison}

\begin{figure}[t]
    \centering
    \includegraphics[width=\linewidth]{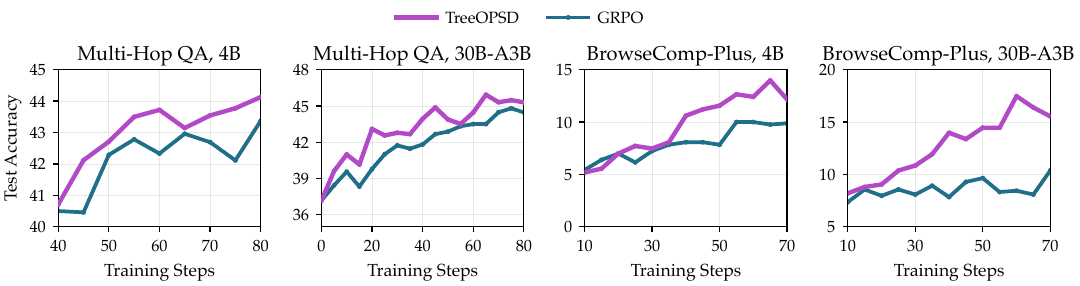}
    \vspace{-10pt}
    \caption{\textbf{Test accuracy during training.} The four panels report TreeOPSD and GRPO with Qwen3-4B-Thinking-2507 and Qwen3-30B-A3B-Thinking-2507 on the Multi-Hop QA average (HotpotQA, 2WikiMultiHopQA, MuSiQue, and Bamboogle) and on BrowseComp-Plus. Higher curves indicate stronger final policies under identical sampling budgets.}
    \label{fig:training-curves}
\end{figure}

Table~\ref{tab:main-results} and Figure~\ref{fig:training-curves} show TreeOPSD with the highest Multi-Hop QA and Deep Research averages at both scales. TreePO is trained on the same initialize-then-expand trees, the same credits $V_i$, and the same optimizer blocks, so the two methods differ only in how that credit is used. TreePO keeps the collection-time coefficient $V_i/L_x$ on every later block. TreeOPSD matches a frozen turn teacher and recomputes the residual as $\pi_\theta$ moves. TreeOPSD improves both Multi-Hop QA and Deep Research, with a larger gain on the longer-horizon Deep Research setting, where it raises the average over TreePO from 13.07 to 15.45 for Qwen3-4B and from 17.59 to 21.20 for Qwen3-30B-A3B. On that longer horizon, later blocks more often update a policy that has already left $\pi_{\theta_{\mathrm{old}}}$. A finite teacher can stop or reverse at that point, while a fixed credit keeps the original sign. The same policies remain ahead on BrowseComp, GAIA, and xbench, which use open-web search rather than the training corpus. Appendix~\ref{app:interaction-horizon} also shows the performance gain under the same training time.

\begin{table}[t]
  \centering
  \caption{\textbf{Performance comparison on Multi-Hop QA and Deep Research.} We train on HotpotQA and OpenResearcher, and evaluate Qwen3-4B-Thinking-2507 and Qwen3-30B-A3B-Thinking-2507 on four test sets for each scenario. TreeOPSD consistently outperforms competitive baselines on average. The best results are indicated in \textbf{bold}.}
  \label{tab:main-results}
  \renewcommand\arraystretch{0.96}
  \setlength{\tabcolsep}{1.05mm}
  \footnotesize
  \vspace{+2pt}
  \begin{tabular}{cccccc|ccccc}
      \toprule
      \multirow{2}{*}{\textbf{Method}} &\multicolumn{5}{c}{\textbf{Multi-Hop QA}} &\multicolumn{5}{c}{\textbf{Deep Research}} \\
      \cmidrule(lr){2-6} \cmidrule(lr){7-11}
      &\textbf{Hotpot} &\textbf{2Wiki} &\textbf{MuSiQue} &\textbf{Bamboogle} &\textbf{Avg} &\textbf{BC+} &\textbf{BC} &\textbf{GAIA} &\textbf{xbench} &\textbf{Avg} \\
      \midrule
      \multicolumn{11}{c}{\textbf{Qwen3-4B-Thinking-2507}} \\
      \cmidrule(lr){1-11}
      \textbf{ReAct}  &31.61 &29.63 &9.68  &25.60 &24.13 &4.70  &1.97 &15.53 &13.00 &8.80 \\
      \textbf{GRPO}   &46.93 &48.62 &21.93 &56.00 &43.37 &10.00 &2.29 &18.45 &18.00 &12.18 \\
      \textbf{GSPO}   &46.55 &49.10 &21.76 &55.20 &43.15 &9.88  &2.37 &18.45 &18.00 &12.17 \\
      \textbf{TreePO} &47.25 &49.90 &22.05 &54.40 &43.40 &11.08 &2.76 &19.42 &19.00 &13.07 \\
      \cellcolor{gray!20}\textbf{TreeOPSD}
       &\cellcolor{gray!20}48.16
       &\cellcolor{gray!20}52.00
       &\cellcolor{gray!20}22.76
       &\cellcolor{gray!20}53.60
       &\cellcolor{gray!20}\textbf{44.13}
       &\cellcolor{gray!20}13.98
       &\cellcolor{gray!20}3.48
       &\cellcolor{gray!20}21.36
       &\cellcolor{gray!20}23.00
       &\cellcolor{gray!20}\textbf{15.45} \\
      \midrule
      \multicolumn{11}{c}{\textbf{Qwen3-30B-A3B-Thinking-2507}} \\
      \cmidrule(lr){1-11}
      \textbf{ReAct}  &41.49 &42.89 &17.87 &46.40 &37.16 &7.23  &3.63 &30.10 &18.00 &14.74 \\
      \textbf{GRPO}   &47.25 &52.13 &23.83 &56.00 &44.80 &10.36 &4.50 &31.07 &19.00 &16.23 \\
      \textbf{GSPO}   &47.70 &52.50 &24.00 &55.20 &44.85 &11.08 &4.74 &32.04 &18.00 &16.47 \\
      \textbf{TreePO} &48.10 &51.80 &24.29 &56.00 &45.05 &12.29 &5.06 &33.01 &20.00 &17.59 \\
      \cellcolor{gray!20}\textbf{TreeOPSD}
       &\cellcolor{gray!20}48.89
       &\cellcolor{gray!20}50.88
       &\cellcolor{gray!20}24.74
       &\cellcolor{gray!20}59.20
       &\cellcolor{gray!20}\textbf{45.93}
       &\cellcolor{gray!20}17.47
       &\cellcolor{gray!20}6.40
       &\cellcolor{gray!20}34.95
       &\cellcolor{gray!20}26.00
       &\cellcolor{gray!20}\textbf{21.20} \\
      \bottomrule
\end{tabular}
\end{table}

\begin{figure}[t]
    \centering
    \includegraphics[width=\linewidth]{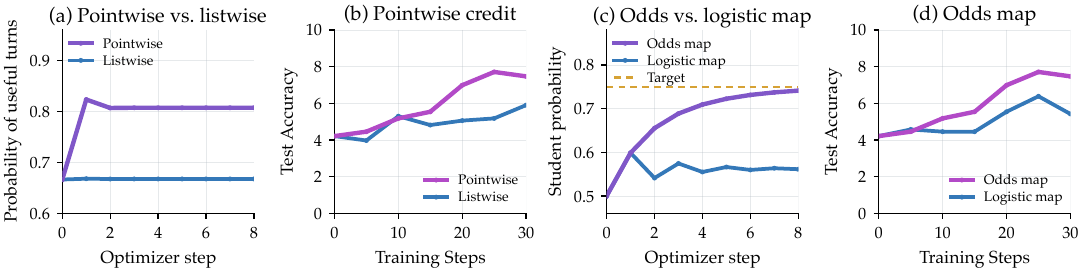}
    \vspace{-10pt}
    \caption{\textbf{Pointwise credit and the posterior map.} (a) Two useful children share a parent backup. Pointwise credit is $V_c$, and listwise credit is the group-relative $V_c-V_i$. (b) BrowseComp-Plus with Qwen3-4B-Thinking-2507. The pointwise curve is the TreeOPSD run from Figure~\ref{fig:training-curves}(c). (c) The odds map and a scaled logistic map agree at the first update, and the student then stops at the odds teacher. (d) The odds map against the logistic map on that split. In (b) and (d), higher curves indicate stronger final policies under identical sampling budgets.}
    \label{fig:target-geometry}
\end{figure}

\subsection{Pointwise credit and the posterior map}
\label{sec:target-geometry}

\paragraph{Pointwise and listwise credit.}
Listwise updates center credit inside the sampled group, which removes the value shared by every child. TreeOPSD uses that child backup $V_c$ as pointwise turn credit, so two useful children can both rise relative to the behavior policy. Figure~\ref{fig:target-geometry} keeps the initialize-then-expand trees and optimizer blocks of TreeOPSD. Panels (a) and (b) replace $V_c$ with the group-relative credit $V_c-V_i$. These centered credits average to zero under the backup weights, so the same two useful children can receive opposite updates. Panel (a) has a parent with two useful children. Matching $V_c$ raises both, while $V_c-V_i$ leaves their total probability near the behavior policy. Panel (b) trains both on BrowseComp-Plus with Qwen3-4B-Thinking-2507, and the pointwise curve lies above the listwise control.
\paragraph{Odds and logistic teachers.}
Panels (c) and (d) keep $V_c$ and change how it becomes a Bernoulli teacher. Eq.~\ref{eq:turn-distribution} uses the odds map $p=r/(1+r)$, and the matching teacher $q=1/2+V/\tau$ is affine in $V$. A parent teacher is then the weighted mean of its child teachers, so tree backup and target construction commute. The logistic map $q=\sigma(V/\tau')$ is also a Bernoulli teacher. Its scale $\tau'$ can match the first tree-policy-gradient update, but its parent teacher is then not the weighted mean of its child teachers. Appendix~\ref{app:mechanistic-numerical} plots the probability of useful turns against their shared credit $m$, and the parent-gradient error when the teacher is not affine in $V$. Panel (d) trains both on BrowseComp-Plus, and the odds map lies above the logistic control.

\begin{table}[t]
    \centering
    \caption{\textbf{Rollout allocation and target-radius sensitivity.} Unweighted four-benchmark averages on Multi-Hop QA and Deep Research with Qwen3-4B-Thinking-2507. Panel (a) varies the initialize-then-expand split at $\rho=0.25$, and panel (b) varies the target radius at $(M,N)=(4,2)$. The gray row is the default setting used in Table~\ref{tab:main-results} and shared by both panels.}
    \label{tab:allocation-radius-sensitivity}
    \renewcommand\arraystretch{1.08}
    \footnotesize
    \vspace{+4pt}
    \begin{tabularx}{\linewidth}{@{}cYYcYY@{}}
        \toprule
        \multicolumn{3}{c}{\textbf{(a) Rollout allocation}} &
        \multicolumn{3}{c}{\textbf{(b) Target radius}} \\
        \cmidrule(r){1-3}\cmidrule(l){4-6}
        $(M,N)$ & \textbf{Multi-Hop QA} & \textbf{Deep Research} &
        $\rho$ & \textbf{Multi-Hop QA} & \textbf{Deep Research} \\
        \midrule
        $(8,0)$ & 43.66 & 11.28 &
        $0.10$ & 43.79 & 14.62 \\
        $(4,0)$ & 42.98 & 10.05 &
        $0.20$ & 44.01 & 15.06 \\
        \rowcolor{gray!20}
        $(4,2)$ & 44.13 & 15.45 &
        $0.25$ & 44.13 & 15.45 \\
        $(2,6)$ & 44.07 & 14.97 &
        $0.30$ & 43.95 & 14.88 \\
        $(2,2)$ & 43.25 & 11.33 &
        $0.40$ & 43.81 & 14.69 \\
        \bottomrule
    \end{tabularx}
\end{table}

\subsection{Rollout allocation and target-radius sensitivity}
\label{sec:allocation-sensitivity}

\paragraph{Rollout allocation.}
Table~\ref{tab:allocation-radius-sensitivity}(a) varies the initialize-then-expand split at the default target radius, comparing independent trajectories with prefix continuations from visited histories. The rows $(8,0)$, $(4,2)$, and $(2,6)$ match GRPO's expected sampling cost, and $(4,0)$ and $(2,2)$ use half of that budget. TreeOPSD recovers tree policy gradient at the behavior policy, so the flat $(8,0)$ tree is the closest first-order control to GRPO, and the two methods stay close on Multi-Hop QA. Moving the same budget onto prefix continuations at $(4,2)$ raises both averages, and it also outperforms the more continuation-heavy $(2,6)$ split. At half budget, $(2,2)$ improves over flat $(4,0)$, and on Deep Research it matches full-budget $(8,0)$, so a moderate number of prefix continuations is a better use of that budget than extra independent trajectories or a continuation-heavy split.
\paragraph{Target radius.}
Table~\ref{tab:allocation-radius-sensitivity}(b) keeps the default tree and varies $\rho$, which sets how far each frozen Bernoulli teacher may move from $1/2$. The first optimizer block does not depend on $\rho$, because the distillation gradient recovers tree policy gradient at the behavior policy. Later blocks close the gap to a teacher whose distance from $\tfrac12$ is set by $\rho$, so the sweep measures those later updates. Both averages peak at the default radius, and Deep Research moves more with $\rho$ than Multi-Hop QA, but accuracy remains close to that peak across the full radius sweep.

\section{Conclusion}
\label{sec:conclusion}

We presented TreeOPSD, a critic-free posterior self-distillation framework for long-horizon agentic RLVR. It backs sparse terminal rewards into implicit turn-level credit on a shared-prefix tree and mitigates policy drift by matching a finite behavior-relative teacher with Forward KL. The affine target commutes with tree backup, and the distillation gradient recovers tree policy gradient at the behavior policy, then stops or reverses after overshoot, so a rollout batch supports sequential updates without a learned critic, process labels, or an external teacher. TreeOPSD improves over matched-budget baselines at both scales on Multi-Hop QA and Deep Research, and the gain is much larger on the longer-horizon setting.

\bibliography{iclr2027_conference}
\bibliographystyle{iclr2025_conference}

\clearpage
\section*{Contents}

\begingroup
\setlength{\parskip}{1pt}
\setlength{\parindent}{0pt}
\newcommand{\treeflowtocline}[2]{%
    \par\addvspace{5pt}%
    \noindent{\bfseries\hyperref[#1]{#2}\hfill\hyperref[#1]{\pageref*{#1}}}\par}
\newcommand{\treeflowtocsubline}[2]{%
    \noindent\hspace*{1.6em}\hyperref[#1]{#2}\dotfill
    \hyperref[#1]{\pageref*{#1}}\par}

\treeflowtocline{sec:introduction}{1\quad Introduction}
\treeflowtocline{sec:preliminaries}{2\quad Preliminaries}
\treeflowtocsubline{sec:agentic-rl}{2.1\quad Agentic reinforcement learning}
\treeflowtocsubline{sec:rollout-trees}{2.2\quad Initialize-then-expand rollout trees}
\treeflowtocsubline{sec:tree-credit}{2.3\quad Tree-based turn credit}
\treeflowtocline{sec:rl-to-tree-projection}{3\quad TreeOPSD}
\treeflowtocsubline{sec:target-and-projection}{3.1\quad Behavior-relative turn distributions}
\treeflowtocsubline{sec:soft-values}{3.2\quad Teacher construction}
\treeflowtocsubline{sec:distribution-matching}{3.3\quad Distribution matching and off-policy analysis}
\treeflowtocsubline{sec:sequential-optimization}{3.4\quad Practical optimization}
\treeflowtocline{sec:experiments}{4\quad Experiments}
\treeflowtocsubline{sec:experimental-setup}{4.1\quad Experimental setup}
\treeflowtocsubline{sec:main-comparison}{4.2\quad Main results}
\treeflowtocsubline{sec:target-geometry}{4.3\quad Pointwise credit and the posterior map}
\treeflowtocsubline{sec:allocation-sensitivity}{4.4\quad Rollout allocation and target-radius sensitivity}
\treeflowtocline{sec:conclusion}{5\quad Conclusion}

\vspace{8pt}
\noindent\textbf{Appendices}\par
\treeflowtocline{app:related-work}{A\quad Related Work}
\treeflowtocline{app:theory}{B\quad Theoretical Analysis}
\treeflowtocsubline{app:credit-details}{B.1\quad Calibrated affine targets on a fixed tree}
\treeflowtocsubline{app:proof-barycenter}{B.2\quad Exact Forward-KL aggregation}
\treeflowtocsubline{app:proof-tangent}{B.3\quad On-policy tangent and off-policy drift analysis}
\treeflowtocline{app:discussion}{C\quad Discussion}
\treeflowtocsubline{app:discussion-layers}{C.1\quad Tree acquisition, credit assignment, and policy optimization}
\treeflowtocsubline{app:discussion-coordinates}{C.2\quad Independent turn targets and pairwise preferences}
\treeflowtocsubline{app:discussion-feedback}{C.3\quad Finite-target feedback under sequential optimization}
\treeflowtocsubline{app:discussion-limitations}{C.4\quad Limitations and future work}
\treeflowtocline{app:extended-results}{D\quad Extended experimental results}
\treeflowtocsubline{app:small-backbones}{D.1\quad Applicability to other model families}
\treeflowtocsubline{app:mechanistic-numerical}{D.2\quad Controlled numerical studies}
\treeflowtocsubline{app:interaction-horizon}{D.3\quad Interaction length and training efficiency in Deep Research}
\treeflowtocsubline{app:seed-variability}{D.4\quad Stability across training seeds}
\treeflowtocline{app:experimental-details}{E\quad Experimental and implementation details}
\treeflowtocsubline{app:datasets-environments}{E.1\quad Datasets, environments, and rewards}
\treeflowtocsubline{app:models}{E.2\quad Models}
\treeflowtocsubline{app:training-evaluation}{E.3\quad Training and evaluation settings}
\treeflowtocsubline{app:baseline-budget}{E.4\quad Baselines and rollout-budget control}
\treeflowtocline{app:data-examples}{F\quad Data Examples}
\treeflowtocsubline{app:data-hotpotqa}{F.1\quad Multi-Hop QA (HotpotQA)}
\treeflowtocsubline{app:data-openresearcher}{F.2\quad Deep Research (OpenResearcher)}
\endgroup

\clearpage
\appendix
\section{Related Work}
\label{app:related-work}

\paragraph{RLVR and agentic RLVR.}
Reinforcement learning with verifiable rewards (RLVR) \citep{shao2024deepseekmath,deepseekai2025deepseekr1,yu2026dapo} trains language models from automatically checked outcomes and has produced strong reasoning policies in mathematics, coding, and other rule-based tasks. PPO remains a general policy-gradient backbone \citep{schulman2017proximal}, while GRPO removes the learned value model by estimating advantages from response groups \citep{shao2024deepseekmath}. Later variants alter the granularity of importance weighting \citep{zheng2025gspo} or organize comparisons within larger groups \citep{feng2025gigrpo}. RLVR has also moved from complete reasoning responses to agents that interleave language, tools, and environment observations \citep{yao2022react}. In this setting, a single terminal advantage supervises a sequence of search queries, tool calls, and answer decisions \citep{jin2025searchr1,zheng2025stepsearch,dong2026arpo}. TreeOPSD focuses on this turn-structured regime and derives intermediate targets directly from verifiable outcomes rather than introducing a learned critic or process labels.

\paragraph{Tree search and tree-based policy optimization.}
Tree search lets language models compare continuations from a shared history instead of drawing only independent complete trajectories. At inference time, Tree-of-Thought \citep{yao2023treeofthoughts}, language-agent search \citep{zhou2024lats}, and MCTS-style procedures \citep{feng2023alphazero} use branching to improve planning or answer selection. Offline methods turn reasoning trees into supervised or preference data \citep{lai2024stepdpo}, while online methods bring the tree into RL for exploration, value estimation, local comparisons, or rollout allocation \citep{hou2025treerl,ji2026treesearch,yang2025treerpo,zong2026at2po,dong2026arpo,zou2026trace}. These approaches differ in where branches are collected and how sibling outcomes become gradients. TreeOPSD takes a shared-prefix tree as input, backs terminal evidence into turn credit, and converts that credit into a finite target for the current policy. Its contribution lies in optimizing the acquired tree rather than choosing where the tree should branch.

\paragraph{Sample-efficient RLVR and rollout allocation.}
A parallel line of work reduces rollout cost by deciding which examples receive sampling budget. Offline curation fixes the training set before optimization \citep{ye2025limo,li2025limr,wang2025oneexample}. Online methods filter prompts whose rollouts are uninformative \citep{yu2026dapo,bae2026online} or predict prompt difficulty before sampling \citep{zheng2025act,qu2025prompt,gao2025prompt,mao2026dynamics,qu2026small}, much as automated model evaluation predicts test accuracy from unlabeled data \citep{peng2024energy}. POPO replays effective groups from earlier policies in place of zero-variance groups \citep{mao2026rlvr}, and budget allocation varies the number of rollouts per prompt \citep{li2025knapsack,yao2026coba}. In supervised fine-tuning, online batch selection scores samples by utility and diversity \citep{zou2025utility}. TRACE extends allocation from prompt roots to turn-level prefixes with a shared success predictor \citep{zou2026trace}. These methods decide which rollouts to acquire, and TreeOPSD decides how an acquired tree updates the policy, so the two compose without changing the teacher in Section~\ref{sec:soft-values}.

\paragraph{Credit assignment under sparse rewards.}
Credit assignment under delayed rewards is a long-standing RL problem. Temporal-difference learning \citep{sutton1998reinforcement}, potential-based shaping \citep{ng1999policy}, and return decomposition \citep{arjona2019rudder} provide denser signals by estimating state values or redistributing terminal returns. For language-model reasoning, process supervision and process reward models score intermediate steps \citep{uesato2022solving,lightman2024verify,wang2024mathshepherd}, and recent work derives progress signals from Monte Carlo continuations, answer likelihoods, or implicit process objectives \citep{setlur2025rewarding,yuan2024freeprocess}. Long-horizon LLM methods further use trajectory graphs \citep{li2026salt}, vine-style continuations \citep{kazemnejad2025vineppo}, or alternative reward aggregation \citep{cheng2026stop} to localize responsibility within a rollout. Rollout trees \citep{ji2026treesearch,zong2026at2po,yang2025treerpo} provide a related source of counterfactual evidence because sibling suffixes keep the prefix fixed while varying the future. TreeOPSD backs this evidence through the sampled tree and preserves value shared by multiple useful continuations instead of removing it through sibling centering.

\paragraph{Behavior-relative targets and posterior distillation.}
KL-regularized RL \citep{ouyang2022training}, maximum-a-posteriori policy optimization \citep{abdolmaleki2018maximum}, and preference optimization \citep{rafailov2023direct,meng2024simpo} improve a policy while anchoring it to a reference distribution. Mirror-descent and mirror-ascent analyses make the corresponding target-and-projection structure explicit \citep{beck2003mirror,vaswani2022general,asad2025fast}, and recent work applies this view to group-based language-model updates on finite response sets \citep{qu2026listwise}. On-policy distillation instead trains on student-generated samples using a teacher distribution \citep{agarwal2024onpolicy,gu2024minillm,zhao2026selfdistilled,kaur2026rethinkingopsd} or privileged continuation information \citep{nguyen2026privileged}. TreeOPSD joins these perspectives at the turn level. It represents each observed turn by a behavior-relative Bernoulli, related to relative density-ratio estimation \citep{yamada2013relative,cranmer2015approximating} and contrastive source classification \citep{goodfellow2014generative,chen2024noise}, then obtains a closed-form teacher from backed-up credit and matches it with Forward KL. Independent turn targets avoid treating a sparse sibling sample as a closed continuation simplex, while the finite teacher supplies a destination that fixed-credit policy gradients do not specify.

\section{Theoretical Analysis}
\label{app:theory}

\subsection{Calibrated affine targets on a fixed tree}
\label{app:credit-details}

We condition on one prompt, its rollout tree, and the frozen behavior policy that sampled that tree. The terminal advantages $A_k$, backed-up turn credits $V_i$, and child weights $\omega_{i,c}$ are fixed during the policy update. For $K\geq2$ terminal trajectories and target radius $0<\rho<1/2$, define
\begin{equation}
    B_K:=\frac{K-1}{\sqrt K},
    \qquad
    \tau:=\frac{B_K}{\rho},
    \qquad
    q_i:=\frac12+\frac{V_i}{\tau},
    \qquad
    \phi_i:=\operatorname{logit}(q_i).
    \label{eq:appendix-notation}
\end{equation}

\paragraph{Lemma B.1 (Depth-independent bound).}
For every terminal occurrence $k$ and every backed-up turn $i$,
\begin{equation}
    |A_k|\leq B_K,
    \qquad
    |V_i|\leq B_K,
    \qquad
    \left|q_i-\frac12\right|\leq\rho.
    \label{eq:depth-independent-bound}
\end{equation}

\paragraph{Proof.}
Let $\xi_k:=R_k-\overline R$ and let $\widehat\sigma_R$ be the sample standard deviation of the $K$ terminal rewards. When $\widehat\sigma_R>0$, we have $\sum_k\xi_k=0$ and $\sum_k\xi_k^2=(K-1)\widehat\sigma_R^2$. Hence
\begin{align}
    \xi_k^2
    &=\left(\sum_{\ell\neq k}\xi_\ell\right)^2
    \leq(K-1)\sum_{\ell\neq k}\xi_\ell^2
    \nonumber\\
    &=(K-1)\left((K-1)\widehat\sigma_R^2-\xi_k^2\right),
\end{align}
which gives $|A_k|=|\xi_k|/\widehat\sigma_R\leq(K-1)/\sqrt K=B_K$. This is Samuelson's inequality \citep{samuelson1968deviant}. Equality holds when one reward differs from the other $K-1$ rewards, which are equal. If $\widehat\sigma_R=0$, every $A_k$ is zero by convention. Because each internal credit is a convex combination of its children, $|V_i|\leq B_K$, and Eq.~\ref{eq:appendix-notation} gives the target-radius bound. \hfill$\square$

\paragraph{Backup-consistent target map.}
Suppose a map $f:[-B_K,B_K]\to(0,1)$ commutes with every finite convex combination, satisfies $f(-v)=1-f(v)$, and maps $B_K$ to $1/2+\rho$. Symmetry then maps $-B_K$ to $1/2-\rho$. Writing any $v\in[-B_K,B_K]$ as the convex interpolation
\begin{equation}
    v=(1-t)(-B_K)+tB_K,
    \qquad
    t=\frac{v+B_K}{2B_K},
\end{equation}
then gives
\begin{equation}
    f(v)=\frac12+\frac{\rho}{B_K}v
    =\frac12+\frac{v}{\tau}.
    \label{eq:unique-affine-solution}
\end{equation}
Thus the affine teacher is the unique symmetric map with this radius. Applying it before or after any normalized tree backup leaves the parent target unchanged.

\paragraph{Proposition B.2 (Trust-region properties and closed-form teacher).}
Parameterize any Bernoulli distribution $\operatorname{Ber}(q)$ by its log-odds $s=\log\frac{q}{1-q}$, so that $q=\sigma(s)$ and $1-q=\sigma(-s)$. The divergence from the on-policy behavior anchor $\operatorname{Ber}(1/2)$ satisfies
\begin{equation}
    \psi(s)
    :=D_{\mathrm{KL}}\!\left(\operatorname{Ber}(\tfrac12)\Vert\operatorname{Ber}(\sigma(s))\right)
    =-\frac12\log\bigl(4\sigma(s)\sigma(-s)\bigr)
    =\log\cosh\frac{s}{2}.
\end{equation}
The penalty $\psi(s)$ has three structural properties:
(i)~\emph{Local quadratic trust region:} $\psi(0)=\psi'(0)=0$ and $\psi''(s)=\sigma(s)(1-\sigma(s))\in(0,1/4]$, giving the Taylor expansion $\psi(s)=\frac{1}{8}s^2-\frac{1}{192}s^4+\mathcal{O}(s^6)$ around the on-policy point $s=0$.
(ii)~\emph{Globally bounded slope:} Its derivative $\psi'(s)=\frac12\tanh\frac{s}{2}=\sigma(s)-\frac12=q-\frac12\in(-1/2,1/2)$ is linear in $q$ and uniformly bounded for all $s\in\mathbb{R}$.
(iii)~\emph{Unique affine maximizer:} Substituting $\psi(s)$ into Eq.~\ref{eq:bernoulli-target} gives the strictly concave objective $\mathcal{J}_i(s)=V_i s-\tau\log\cosh(s/2)$ with derivatives
\begin{equation}
    \mathcal{J}_i'(s)
    =V_i-\tau\left(\sigma(s)-\frac12\right),
    \qquad
    \mathcal{J}_i''(s)
    =-\tau\sigma(s)(1-\sigma(s))<0.
\end{equation}
Because $\psi'(s)=q-1/2$ is linear in $q$, the stationary condition $\mathcal{J}_i'(s_i^\star)=0$ yields the exact affine teacher $\sigma(s_i^\star)=1/2+V_i/\tau=q_i$ required by Eq.~\ref{eq:unique-affine-solution} (whereas the exponential-tilting objective $V_iq-\tau D_{\mathrm{KL}}(\operatorname{Ber}(q)\Vert\operatorname{Ber}(\tfrac12))$ yields a nonlinear logistic teacher $q=\sigma(V_i/\tau)$ that fails to commute with tree backup). By Lemma~B.1 and $\rho<1/2$, we have $\tau>2|V_i|$, so $s_i^\star=\phi_i$ is interior and finite, with target likelihood ratio
\begin{equation}
    r_i^\star=e^{s_i^\star}
    =\frac{q_i}{1-q_i}
    =\frac{\tau+2V_i}{\tau-2V_i}
    \in
    \left[
        \frac{1-2\rho}{1+2\rho},
        \frac{1+2\rho}{1-2\rho}
    \right].
    \qquad\square
\end{equation}

\subsection{Exact Forward-KL aggregation}
\label{app:proof-barycenter}

Let $\mathcal{D}_i$ be the terminal descendants of turn $i$. For a descendant $u$, let $\alpha_{i,u}$ be the product of the normalized child weights along the unique path from $i$ to $u$. If $\mathcal{U}$ is any complete cut below $i$, repeated application of the backup in Eq.~\ref{eq:policy-backup} gives
\begin{equation}
    \sum_{u\in\mathcal{U}}\alpha_{i,u}=1,
    \qquad
    V_i=\sum_{u\in\mathcal{U}}\alpha_{i,u}V_u,
    \qquad
    q_i=\sum_{u\in\mathcal{U}}\alpha_{i,u}q_u.
    \label{eq:affine-cut-consistency}
\end{equation}
The last identity follows from the affine credit-to-target map proved above.

\paragraph{Proof of Theorem~\ref{thm:exact-marginalization}.}
Write $d_{\mathrm B}(a\Vert b):=D_{\mathrm{KL}}(\operatorname{Ber}(a)\Vert\operatorname{Ber}(b))$ and let $H_{\mathrm B}$ denote binary entropy. When every descendant teacher on the cut is matched to the same ancestor probability $p_{\theta,i}\in(0,1)$,
\begin{align}
    \sum_{u\in\mathcal{U}}\alpha_{i,u}d_{\mathrm B}(q_u\Vert p_{\theta,i})
    &=-\sum_{u\in\mathcal{U}}\alpha_{i,u}H_{\mathrm B}(q_u)
      -q_i\log p_{\theta,i}-(1-q_i)\log(1-p_{\theta,i})
    \nonumber\\
    &=\sum_{u\in\mathcal{U}}\alpha_{i,u}
      d_{\mathrm B}(q_u\Vert q_i)
      +d_{\mathrm B}(q_i\Vert p_{\theta,i}).
    \label{eq:latent-risk-decomposition}
\end{align}
The first term on the right is independent of $\theta$. Therefore
\begin{equation}
    \nabla_\theta
    \sum_{u\in\mathcal{U}}\alpha_{i,u}
    d_{\mathrm B}(q_u\Vert p_{\theta,i})
    =
    \nabla_\theta
    d_{\mathrm B}(q_i\Vert p_{\theta,i}),
    \qquad
    \arg\min_{p\in(0,1)}
    \sum_{u\in\mathcal{U}}\alpha_{i,u}d_{\mathrm B}(q_u\Vert p)
    =q_i.
\end{equation}
Setting $\mathcal{U}=\mathcal{D}_i$ (or $\mathcal{U}=\mathcal{C}_i$) yields Eqs.~\ref{eq:loss-decomposition}--\ref{eq:tree-consistent-teacher}. Hence one backed-up teacher produces the exact same gradient on the shared turn as all descendant teachers, while each descendant turn retains its own policy score and target. \hfill$\square$

\subsection{On-policy tangent and off-policy drift analysis}
\label{app:proof-tangent}

\paragraph{Variational duality and Fenchel--Young form.}
Define the convex log-partition potential and its Fenchel conjugate on $(-\tau/2, \tau/2)$:
\begin{equation}
    \Omega_\tau(s):=\tau\log\left(2\cosh\frac{s}{2}\right),
    \qquad
    \Omega_\tau^\star(V):=\sup_{s\in\mathbb{R}}\{Vs-\Omega_\tau(s)\}
    =-\tau H_{\mathrm B}\!\left(\frac12+\frac{V}{\tau}\right).
\end{equation}
For $q_i=q(V_i)=1/2+V_i/\tau$ and $p_{\theta,i}=\sigma(s_{\theta,i})$, expanding $\tau d_{\mathrm B}(q_i\Vert\sigma(s_{\theta,i}))$ yields the Fenchel--Young loss identity of \citet{blondel2020fenchel}:
\begin{equation}
    \tau d_{\mathrm B}\bigl(q(V_i)\Vert\sigma(s_{\theta,i})\bigr)
    =\Omega_\tau(s_{\theta,i})+\Omega_\tau^\star(V_i)-V_i s_{\theta,i}
    =\mathcal{J}_i(Q_i^\star)-\mathcal{J}_i(P_{\theta,i}).
    \label{eq:app-fenchel-young}
\end{equation}
Summing Eq.~\ref{eq:app-fenchel-young} across $i\in\mathcal{T}_x$ with weights $L_i/L_x$ proves the second equality in Eq.~\ref{eq:treeopsd-loss}. Moreover, defining the tree-weighted proximal RL objective $\mathcal{J}_x(\theta):=\frac{1}{L_x}\sum_{i\in\mathcal{T}_x}L_i\mathcal{J}_i(P_{\theta,i})$, Eq.~\ref{eq:app-fenchel-young} implies that for any two parameter iterates $\theta,\theta'$ on a fixed rollout batch,
\begin{equation}
    \mathcal{J}_x(\theta')-\mathcal{J}_x(\theta)
    =\mathcal{L}_x(\theta)-\mathcal{L}_x(\theta').
    \label{eq:monotonic-proximal-improvement}
\end{equation}
Thus every sequential mini-batch step that reduces the Forward-KL distillation loss $\mathcal{L}_x$ monotonically improves the proximal objective $\mathcal{J}_x$ by the exact same amount.

\paragraph{Proof of Theorem~\ref{thm:pg-tangent}.}
For token $j$ in turn $i$, differentiating $s_{\theta,i}$ in Eq.~\ref{eq:turn-score} contributes a factor $1/L_i$, which cancels the turn-length weight $L_i$ in Eq.~\ref{eq:treeopsd-loss}:
\begin{align}
    \frac{\partial\mathcal{L}_x}
    {\partial\log\pi_\theta(a_{i,j}\mid h_{i,j})}
    &=\frac{\tau L_i}{L_x}
      \frac{\partial d_{\mathrm B}(q_i\Vert\sigma(s_{\theta,i}))}{\partial s_{\theta,i}}
      \frac{\partial s_{\theta,i}}{\partial\log\pi_\theta(a_{i,j}\mid h_{i,j})}
    =\frac{\tau L_i}{L_x}
      \bigl(p_{\theta,i}-q_i\bigr)\frac1{L_i}
    \nonumber\\
    &=-\frac1{L_x}
      \left[V_i-\tau\left(p_{\theta,i}-\frac12\right)\right].
    \label{eq:token-projection-gradient}
\end{align}
Define the off-policy effective ascent weight
\begin{equation}
    \Gamma_i(s)
    :=\tau\bigl(q_i-\sigma(s)\bigr)
    =V_i-\tau\left(\sigma(s)-\frac12\right)
    =V_i-\frac{\tau}{2}\tanh\frac{s}{2}.
    \label{eq:bernoulli-score-gradient}
\end{equation}
Substituting $q_i=r_i^\star/(1+r_i^\star)$ and $\sigma(s_{\theta,i})=r_{\theta,i}/(1+r_{\theta,i})$ gives the self-normalized ratio form in Eq.~\ref{eq:effective-credit}:
\begin{equation}
    \Gamma_i(s_{\theta,i})
    =\tau\left(\frac{r_i^\star}{1+r_i^\star}-\frac{r_{\theta,i}}{1+r_{\theta,i}}\right)
    =\tau\frac{r_i^\star-r_{\theta,i}}{(1+r_i^\star)(1+r_{\theta,i})}.
\end{equation}
The three properties in Theorem~\ref{thm:pg-tangent} follow directly:
(i)~At $\theta=\theta_{\mathrm{old}}$, $s_{\theta,i}=0$ and $\Gamma_i(0)=V_i$, so
\begin{equation}
    \left.-\nabla_\theta\mathcal{L}_x(\theta)\right|_{\theta=\theta_{\mathrm{old}}}
    =\frac1{L_x}
      \sum_{i\in\mathcal{T}_x}\sum_{j=1}^{L_i}
      V_i\nabla_\theta
      \log\pi_{\theta_{\mathrm{old}}}(a_{i,j}\mid h_{i,j})
    =g_x^{\mathrm{tree}}.
    \label{eq:behavior-tangent-proof}
\end{equation}
(ii)~Because $\sigma$ is $1/4$-Lipschitz and $\sigma(0)=1/2$, the additive drift correction satisfies
\begin{equation}
    |\Gamma_i(s)-V_i|
    =\tau\left|\sigma(s)-\frac12\right|
    \leq\frac{\tau}{4}|s|,
    \qquad
    |\Gamma_i(s)|
    =\tau|q_i-\sigma(s)|
    \leq\left(\frac12+\rho\right)\tau,
    \label{eq:behavior-drift-response}
\end{equation}
where the uniform bound uses $q_i\in[1/2-\rho,1/2+\rho]$ and $\sigma(s)\in(0,1)$. Unlike unclipped importance-weighted coefficients $r_{\theta,i}V_i=e^{s_{\theta,i}}V_i$, which grow exponentially when $s_{\theta,i}>0$, $\Gamma_i(s_{\theta,i})$ remains bounded for all $r_{\theta,i}\in(0,\infty)$.
(iii)~Because $\sigma$ is strictly increasing and $q_i=\sigma(\phi_i)$,
\begin{equation}
    \Gamma_i(0)=V_i,
    \qquad
    \Gamma_i(\phi_i)=0,
    \qquad
    \operatorname{sign}\Gamma_i(s)=\operatorname{sign}(\phi_i-s)=\operatorname{sign}(r_i^\star-r_{\theta,i}).
    \qquad\square
    \label{eq:residual-sign}
\end{equation}

\paragraph{Off-policy scalar contraction and parameter-drift bound.}
By the mean-value theorem, for any score $s\in\mathbb{R}$ there exists $\xi$ between $s$ and $\phi_i$ such that $\Gamma_i(s)=\tau\sigma'(\xi)(\phi_i-s)$ with $0<\sigma'(\xi)\leq 1/4$. A gradient step $s^{(t+1)}=s^{(t)}+\eta\frac{L_i}{L_x}\Gamma_i(s^{(t)})$ on the turn score therefore satisfies
\begin{equation}
    s^{(t+1)}-\phi_i
    =\bigl(1-\kappa_t\bigr)\bigl(s^{(t)}-\phi_i\bigr),
    \qquad
    \kappa_t:=\eta\frac{\tau L_i}{L_x}\sigma'(\xi_t)\in\left(0,\frac{\eta\tau L_i}{4L_x}\right].
    \label{eq:scalar-contraction}
\end{equation}
Whenever $0<\eta\leq\frac{4L_x}{\tau L_i}$, we have $\kappa_t\in(0,1]$, so $s^{(t)}$ contracts monotonically toward the target log-odds $\phi_i=\log r_i^\star$ without overshooting. Finally, in parameter space, let $\boldsymbol d_i(\theta):=\sum_{j=1}^{L_i}\nabla_\theta\log\pi_\theta(a_{i,j}\mid h_{i,j})$ and compare the TreeOPSD update direction $\boldsymbol g_x^{\mathrm{OPSD}}(\theta)$ with the fixed-credit direction $\boldsymbol g_x^{\mathrm{fixed}}(\theta)$:
\begin{align}
    \boldsymbol g_x^{\mathrm{OPSD}}(\theta)
    &:=
    \frac1{L_x}\sum_{i\in\mathcal{T}_x}
    \Gamma_i(s_{\theta,i})\boldsymbol d_i(\theta),
    &
    \boldsymbol g_x^{\mathrm{fixed}}(\theta)
    &:=
    \frac1{L_x}\sum_{i\in\mathcal{T}_x}
    V_i\boldsymbol d_i(\theta).
\end{align}
Applying Eq.~\ref{eq:behavior-drift-response} gives the off-policy parameter-drift bound
\begin{equation}
    \left\|
    \boldsymbol g_x^{\mathrm{OPSD}}(\theta)
    -\boldsymbol g_x^{\mathrm{fixed}}(\theta)
    \right\|
    \leq
    \frac{\tau}{4L_x}
    \sum_{i\in\mathcal{T}_x}
    |s_{\theta,i}|\,\|\boldsymbol d_i(\theta)\|.
    \label{eq:parameter-drift-bound}
\end{equation}
Thus TreeOPSD and fixed-credit optimization coincide at the on-policy first mini-batch ($s_{\theta,i}=0$) and depart proportionally to the off-policy score drift $|s_{\theta,i}|$ accumulated across subsequent mini-batches.

\section{Discussion}
\label{app:discussion}

\subsection{Tree acquisition, credit assignment, and policy optimization}
\label{app:discussion-layers}

Tree-based agent RL involves three decisions that are often coupled in one algorithm. A rollout procedure first decides which prompts or prefixes receive continuations. A credit-assignment rule then maps terminal outcomes on that tree to turn values. Policy optimization uses those values to update the current LLM. TreeOPSD handles the last step: it takes a sampled tree and normalized child weights, converts the backed-up values into behavior-relative targets, and matches the current policy to those targets. The same target construction can be paired with random branching, learned rollout allocation, or other tree samplers.

\subsection{Independent turn targets and pairwise preferences}
\label{app:discussion-coordinates}

A categorical distribution over sampled siblings is invariant to a common shift of their logits or rewards. It can rank the observed continuations, but it removes value shared by all of them. This is appropriate when the sampled set is treated as a closed choice set. In an LLM, however, a few sampled siblings cover only a small part of the continuation space. If every observed sibling is useful relative to the behavior policy, a sibling-normalized update cannot raise them together. TreeOPSD instead assigns each observed turn its own behavior-relative Bernoulli target, allowing the shared policy to increase all useful turns while retaining their relative differences. Each target describes where one observed turn should move relative to its behavior likelihood. These targets do not form a normalized action distribution over the full vocabulary or continuation space.

Step-level preference optimization is a boundary case. For two siblings with centered credits $V_+=v$ and $V_-=-v$, the TreeOPSD behavior-policy tangent is proportional to the usual pairwise direction $\nabla\log\pi(a^+)-\nabla\log\pi(a^-)$. The objectives differ after that initial point. A hard pairwise preference keeps enlarging the preferred margin, whereas TreeOPSD assigns both turns finite soft targets and stops when those targets are reached. When the pair has nonzero mean credit, TreeOPSD also retains the common update that pairwise centering removes.

\subsection{Finite-target feedback under sequential optimization}
\label{app:discussion-feedback}

Rollout batches are generated by a frozen behavior policy, but later optimizer blocks run after earlier blocks have already changed the shared LLM parameters. A fixed-credit surrogate continues to apply the collection-time coefficient $V_i$ even when the likelihood of turn $i$ has already moved. TreeOPSD keeps the teacher fixed and recomputes the student probability, replacing $V_i$ by the residual $\Gamma_i=\tau(q_i-p_{\theta,i})$. The first block therefore follows the tree-policy-gradient tangent. Later blocks reduce, stop, or reverse the update according to the current student--teacher gap.

This feedback differs from importance-ratio clipping. Clipping limits how strongly a sampled action can move, but it does not specify a destination. The TreeOPSD teacher supplies that destination in a behavior-relative coordinate. The radius $\rho$ bounds the target probability and, with it, the geometric-mean likelihood ratio of the observed turn. It sets the finite target that later blocks pursue after updates on other prompts have already moved the shared policy.

\subsection{Limitations and future work}
\label{app:discussion-limitations}

\paragraph{Allocation-aware teachers.}
Theorems~\ref{thm:exact-marginalization} and~\ref{thm:pg-tangent} condition on the sampled tree, so TreeOPSD can train on trees from learned allocators such as DPS \citep{mao2026dynamics}, GPS \citep{qu2026small}, and TRACE \citep{zou2026trace}, while its teachers still come only from verifier outcomes and behavior likelihoods. The allocator also sets the scale of those teachers. Under the Samuelson calibration, the root-mean-square of $q_{i_k}-1/2$ over the $K$ leaves equals $\rho/\sqrt{K-1}$ whenever the rewards are not all equal, about $0.075$ at the default $\rho=0.25$ and $K=12$, and only a tree with a single dissenting reward reaches the full radius. Allocators that favor mixed outcomes move trees toward balanced rewards, and budget allocators vary $K$ across prompts \citep{li2025knapsack,yao2026coba}. A radius set from the realized credits or from $K$ would let allocation and teacher scale be designed together. It keeps both theorems, since the first-block tangent does not depend on $\tau$ and Theorem~\ref{thm:exact-marginalization} only needs every $q_i$ in $(0,1)$.

\paragraph{Tree replay.}
Each tree currently serves $E=4$ optimizer blocks before it is discarded. POPO saves rollouts by replaying effective groups from earlier policies in place of zero-variance groups \citep{mao2026rlvr}, and the same replay fits TreeOPSD. A target ratio $r_i^\star$ is defined relative to the policy that collected its tree, so a replayed teacher still satisfies Eq.~\ref{eq:teacher-trust-region}, and the residual $\tau(q_i-p_{\theta,i})$ shrinks or reverses on turns that the current policy has already moved toward or past their targets. The residual also scores stored turns with one forward pass, which could drive utility- and diversity-aware batch selection over a replay buffer \citep{zou2025utility}. Since each $V_i$ averages continuations from the collecting policy, replayed credit ages as the policy improves, and weighting stored trees by their off-policiness, as POPO does for groups, would account for this aging.

\paragraph{Continual agentic post-training.}
A tree whose rewards all coincide has $q_i=1/2$ at every turn, so in later blocks it pulls its own turns back toward $\pi_{\theta_{\mathrm{old}}}$, much as Learning without Forgetting distills the previous model's outputs \citep{li2017learning}. On-policy RL forgets less than SFT, and its forgetting tracks the KL shift on the new task \citep{shenfeld2026rl}. TreeOPSD shares the on-policy sampling behind this result, and its teachers also stay within $\rho$ of $1/2$. RL post-training still forgets across task sequences, however, because KL measured on current-task data does not control drift on prior-task distributions \citep{luo2026rl}. Mixing a few stored trees from earlier environments, together with their logged behavior probabilities, into later batches would extend the $q_i=1/2$ anchor to prior tasks without new rollouts. On the parameterization side, continual-learning methods mitigate the stability--plasticity dilemma through parameter regularization \citep{kirkpatrick2017overcoming}, fly-olfactory-circuit designs \citep{zou2025structural,zou2026flycl}, and guided random projections \citep{li2026enhancing}. FlyLoRA limits inter-task interference in LoRA \citep{hu2022lora} by using a frozen sparse random projection as an implicit router over rank-wise experts \citep{zou2025flylora}, and TreeOPSD applies to such parameterizations unchanged because it only reweights token log-likelihood gradients. The residual already accounts for drift among prompts within a batch (Appendix~\ref{app:discussion-feedback}), and a decorrelated parameterization would limit interference across environments that never share a batch.

\section{Extended experimental results}
\label{app:extended-results}

\subsection{Applicability to other model families}
\label{app:small-backbones}

We train Qwen3-1.7B and Llama-3.2-3B-Instruct on HotpotQA and evaluate them with the same four-benchmark exact-match average as Section~\ref{sec:main-comparison}. The ReAct scaffold, local E5 retrieval server \citep{wang2022text}, and TreeOPSD settings ($(M,N)=(4,2)$, $\rho=0.25$, $E=4$) match the main Multi-Hop QA runs.

\begin{table}[t]
  \centering
  \caption{\textbf{Performance comparison on Multi-Hop QA with smaller backbones.} We train on HotpotQA and evaluate Qwen3-1.7B and Llama-3.2-3B-Instruct on four test sets. TreeOPSD consistently outperforms competitive baselines on average. The best results are indicated in \textbf{bold}.}
  \label{tab:small-backbone-results}
  \renewcommand\arraystretch{0.96}
  \setlength{\tabcolsep}{0.80mm}
  \footnotesize
  \vspace{+4pt}
  \begin{tabular}{cccccc|ccccc}
      \toprule
      \multirow{2}{*}{\textbf{Method}} & \multicolumn{5}{c}{\textbf{Qwen3-1.7B}} & \multicolumn{5}{c}{\textbf{Llama-3.2-3B-Instruct}} \\
      \cmidrule(lr){2-6} \cmidrule(lr){7-11}
      &\textbf{Hotpot} &\textbf{2Wiki} &\textbf{MuSiQue} &\textbf{Bamboogle} &\textbf{Avg} &\textbf{Hotpot} &\textbf{2Wiki} &\textbf{MuSiQue} &\textbf{Bamboogle} &\textbf{Avg} \\
      \midrule
          \textbf{ReAct}     &23.04 &26.11 &6.37  &19.20 &18.68 &12.18 &7.38  &4.22  &20.00 &10.95 \\
          \textbf{GRPO}      &29.10 &31.48 &8.61  &33.60 &25.70 &36.80 &36.35 &13.78 &40.00 &31.73 \\
          \textbf{GSPO}      &29.09 &31.85 &8.52  &32.80 &25.57 &36.48 &36.73 &13.61 &39.20 &31.50 \\
          \textbf{TreePO}    &29.86 &32.18 &8.90  &33.60 &26.13 &37.56 &37.43 &14.23 &40.00 &32.30 \\
          \cmidrule(lr){1-11}
          \cellcolor{gray!20}\textbf{TreeOPSD}
           &\cellcolor{gray!20}31.30
           &\cellcolor{gray!20}33.81
           &\cellcolor{gray!20}9.60
           &\cellcolor{gray!20}32.80
           &\cellcolor{gray!20}\textbf{26.88}
           &\cellcolor{gray!20}39.59
           &\cellcolor{gray!20}41.15
           &\cellcolor{gray!20}15.97
           &\cellcolor{gray!20}45.60
           &\cellcolor{gray!20}\textbf{35.58} \\
      \bottomrule
\end{tabular}
\end{table}

\begin{figure}[t]
    \vspace{-4pt}
    \centering
    \includegraphics[width=\linewidth]{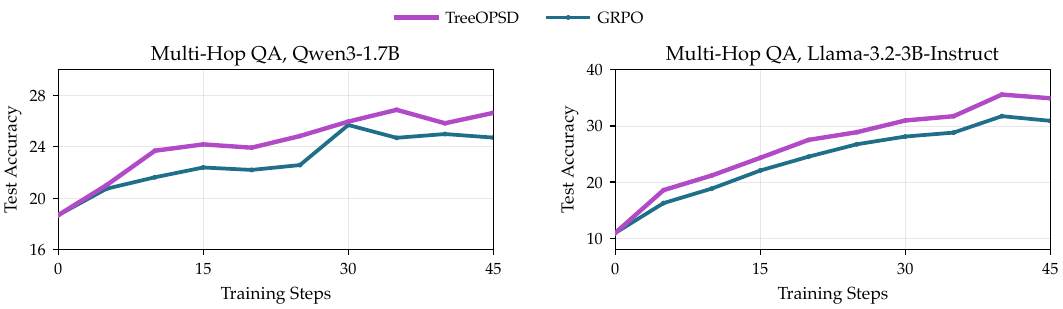}
    \vspace{-10pt}
    \caption{\textbf{Test accuracy during training.} The two panels report TreeOPSD and GRPO with Qwen3-1.7B and Llama-3.2-3B-Instruct on the Multi-Hop QA average (HotpotQA, 2WikiMultiHopQA, MuSiQue, and Bamboogle). Higher curves indicate stronger final policies under identical sampling budgets.}
    \label{fig:small-backbone-training-curves}
\end{figure}

Table~\ref{tab:small-backbone-results} and Figure~\ref{fig:small-backbone-training-curves} show TreeOPSD with the highest average on both backbones. Because TreePO shares the trees, credits $V_i$, and optimizer blocks of TreeOPSD, the remaining gap is the difference between residual matching and fixed credit. The 1.7B TreeOPSD curve lies above GRPO after the first few updates. Llama-3.2-3B-Instruct starts from a weaker ReAct initialization, and TreeOPSD remains ahead of GRPO through the rest of training.

\subsection{Controlled numerical studies}
\label{app:mechanistic-numerical}

\begin{figure}[t]
    \centering
    \includegraphics[width=\linewidth]{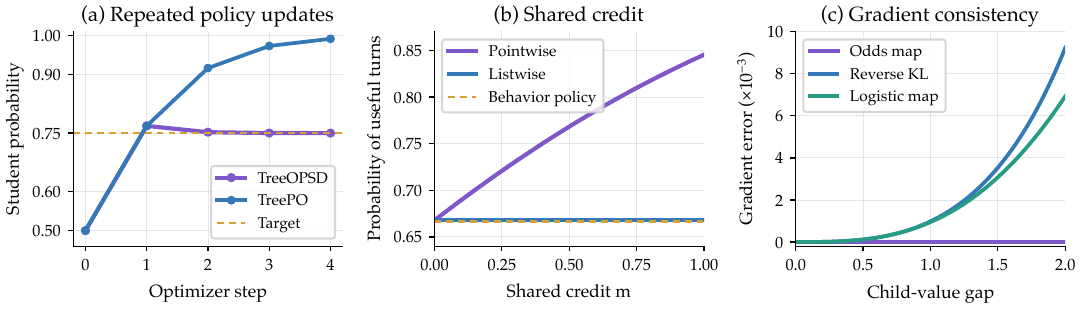}
    \vspace{-10pt}
    \caption{\textbf{Numerical studies of turn credit and the posterior map.} (a) Repeated updates share one scalar policy score. TreeOPSD stops at the teacher, while TreePO keeps the collection-time coefficient. (b) Two useful children share a credit $m$. Pointwise credit is $V_c$, and listwise credit is the group-relative $V_c-V_i$. (c) The odds map with Forward KL preserves the parent gradient after tree backup. Reverse KL and a logistic map do not.}
    \label{fig:mechanistic-numerical}
\end{figure}

Figure~\ref{fig:mechanistic-numerical} isolates the sequential residual in Theorem~\ref{thm:pg-tangent} and the two design choices in Section~\ref{sec:target-geometry} on a noise-free backup.

\paragraph{Finite-target residual.}
Panel (a) starts at the behavior policy, where TreeOPSD and TreePO share the tree-policy-gradient tangent. Later blocks see a student that has already moved. TreePO keeps the collection-time credit $V$, while TreeOPSD evaluates the current residual $\tau(q-p)$. Once the student passes the teacher, that residual changes sign and the update stops at the finite target.
\paragraph{Pointwise credit and the posterior map.}
Panel (b) gives two useful children a shared credit $m$. Pointwise credit is $V_c$, and listwise credit is the group-relative $V_c-V_i$. These centered credits average to zero under the backup weights, so listwise cannot raise both children relative to the behavior policy. Panel (c) keeps $V_c$ and changes the teacher map. The odds map $q=1/2+V/\tau$ is affine in $V$, so a parent teacher is the weighted mean of its child teachers and Forward KL preserves that parent gradient. A logistic map does not commute with backup, and Reverse KL uses a logit-space centroid instead of that mean.

\subsection{Interaction length and training efficiency in Deep Research}
\label{app:interaction-horizon}

\begin{figure}[t]
    \centering
    \includegraphics[width=\linewidth]{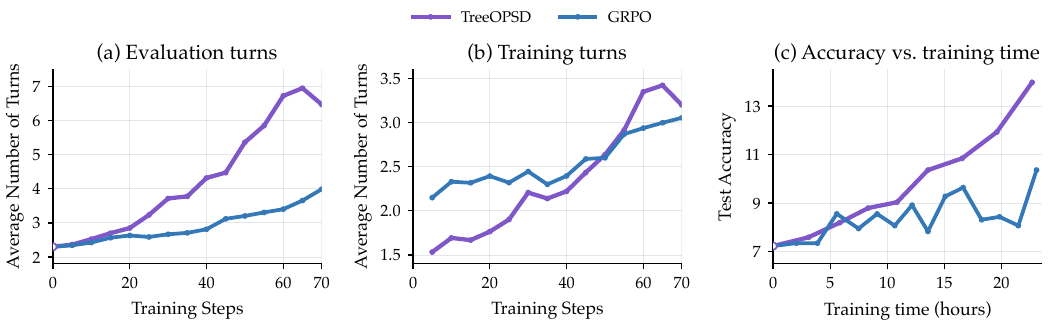}
    \vspace{-10pt}
    \caption{\textbf{Interaction length and training efficiency.} Panels (a) and (b) report the average number of assistant turns during evaluation and training on BrowseComp-Plus with Qwen3-30B-A3B-Thinking-2507. Panel (c) plots test accuracy against cumulative training time and shows TreeOPSD only within the total time used by GRPO. Higher curves in (c) indicate stronger final policies under identical training-time budgets.}
    \label{fig:interaction-horizon}
\end{figure}

Long-horizon search is difficult for outcome-only policy gradients because one terminal advantage supervises every preceding query, document read, and answer decision. Early useful actions in an unsuccessful rollout receive the same trajectory-level signal as the eventual error, whereas branches from a shared prefix show whether the next turn improves the downstream trajectory. Figure~\ref{fig:interaction-horizon} reports this difference with Qwen3-30B-A3B-Thinking-2507 on BrowseComp-Plus. TreeOPSD uses more assistant turns at evaluation, and the training rollouts follow the same trend later. Panel (c) keeps TreeOPSD inside GRPO's measured training-time budget, and the accuracy curve remains above GRPO when training time replaces policy updates on the horizontal axis.

\subsection{Stability across training seeds}
\label{app:seed-variability}

We rerun the Qwen3-4B-Thinking-2507 comparison with five training seeds. Table~\ref{tab:seed-variability} reports mean $\pm$ std of the same Multi-Hop QA and Deep Research averages as Table~\ref{tab:main-results}. ReAct is the frozen initialization, rescored under the same five evaluation draws.

\begin{table}[t]
  \centering
  \caption{\textbf{Mean $\pm$ std over five training seeds with Qwen3-4B-Thinking-2507.} Averages use the same Multi-Hop QA and Deep Research protocol as Table~\ref{tab:main-results}. TreeOPSD consistently outperforms competitive baselines on average. The best results are indicated in \textbf{bold}.}
  \label{tab:seed-variability}
  \renewcommand\arraystretch{1.08}
  \footnotesize
  \vspace{+4pt}
  \begin{tabular}{lcc}
      \toprule
      \textbf{Method} & \textbf{Multi-Hop QA} & \textbf{Deep Research} \\
      \midrule
      \textbf{ReAct}  & $24.13 \pm 0.05$ & $8.80 \pm 0.26$ \\
      \textbf{GRPO}   & $43.37 \pm 0.19$ & $12.18 \pm 0.27$ \\
      \textbf{GSPO}   & $43.15 \pm 0.19$ & $12.17 \pm 0.32$ \\
      \textbf{TreePO} & $43.40 \pm 0.17$ & $13.07 \pm 0.37$ \\
      \cellcolor{gray!20}\textbf{TreeOPSD}
       &\cellcolor{gray!20}$\mathbf{44.13 \pm 0.16}$
       &\cellcolor{gray!20}$\mathbf{15.45 \pm 0.39}$ \\
      \bottomrule
  \end{tabular}
\end{table}

TreePO is the closest trained control because it already uses the sampled tree, the backed-up credits $V_i$, unique-turn reconstruction, which scores each shared turn once, and the four sequential optimizer blocks. Across seeds, the remaining variation comes from which rollouts were collected and which checkpoint in the evaluation window was reported. TreeOPSD shares those sources of noise. The TreeOPSD--TreePO and TreeOPSD--GRPO gaps keep the same order in every seed, and the remaining gap is the difference between residual matching and fixed credit.

The two columns do not scatter the same way. HotpotQA and 2WikiMultiHopQA are large enough that a few extra exact matches barely move the Multi-Hop average. GAIA and xbench-DeepSearch have about one hundred questions each, so one extra match already moves the Deep Research average. ReAct's spread there reflects this granularity.

\section{Experimental and implementation details}
\label{app:experimental-details}

We follow Search-R1 \citep{jin2025searchr1, zou2026trace} for Multi-Hop QA and OpenResearcher \citep{li2026openresearcher, tao2026trace} for Deep Research.

\subsection{Datasets, environments, and rewards}
\label{app:datasets-environments}

\subsubsection{Multi-Hop QA}

\paragraph{Training data and environment.} We use all 90,447 questions from the HotpotQA distractor training split \citep{yang2018hotpotqa}. The agent searches a local E5 retrieval server \citep{wang2022text} built over a Wikipedia dump \citep{karpukhin2020dense}. Each tool call accepts up to four queries and returns the top three passages for each query. We cap a rollout at five assistant turns, including the final answer, which leaves at most four tool-calling turns.

\paragraph{Evaluation benchmarks.} We evaluate on the 7,405-question HotpotQA validation split, the 12,576-question 2WikiMultiHopQA development split \citep{ho2020constructing}, the 2,417-question MuSiQue development split \citep{trivedi2022musique}, and all 125 Bamboogle test questions \citep{press2023measuring}.

\paragraph{Reward function.} Training uses thresholded token F1. A parsed answer receives its token-level F1 when that score is at least $0.3$ and receives zero otherwise. Validation records both F1 and exact match. Table~\ref{tab:main-results} and the training curves use exact match (EM).

\subsubsection{Deep Research}

\paragraph{Training data and environment.} OpenResearcher provides 16 synthesized trajectories for each of 6,102 MiroVerse questions \citep{li2026openresearcher}. We keep one prompt per unique question and do not place the released teacher responses in the policy input. Training and BrowseComp-Plus \citep{chen2025browsecompplus} evaluation use closed-corpus retrieval. BrowseComp \citep{wei2025browsecomp}, GAIA \citep{mialon2024gaia}, and xbench-DeepSearch \citep{chen2025xbench} use the same search, open, and find tools with a Serper backend.

\paragraph{Evaluation benchmarks.} Final evaluation covers all 830 BrowseComp-Plus test questions, 1,266 BrowseComp test questions, 103 text-only questions from the GAIA development split, and 100 xbench-DeepSearch test questions. To limit temporal variation in live search, we interleave the open-web methods within the same evaluation window and archive the raw search responses with each generation.

\paragraph{Reward function.} The training reward is exact match, with a $0.1$ bonus for a well-formed final-answer span. Table~\ref{tab:main-results} reports exact match without that bonus. No evaluation question from either setting is used for policy optimization.

\subsection{Models}
\label{app:models}

The main experiments use Qwen3-4B-Thinking-2507 and Qwen3-30B-A3B-Thinking-2507 \citep{yang2025qwen3}. Appendix~\ref{app:small-backbones} also trains Qwen3-1.7B and Llama-3.2-3B-Instruct. ReAct \citep{yao2022react} evaluates the corresponding initialization in the same agent interface. The model repositories are:
\begin{itemize}[leftmargin=1.6em,itemsep=0pt,topsep=2pt]
    \item Qwen3-4B-Thinking-2507: \url{https://huggingface.co/Qwen/Qwen3-4B-Thinking-2507}.
    \item Qwen3-30B-A3B-Thinking-2507: \url{https://huggingface.co/Qwen/Qwen3-30B-A3B-Thinking-2507}.
    \item Qwen3-1.7B: \url{https://huggingface.co/Qwen/Qwen3-1.7B}.
    \item Llama-3.2-3B-Instruct: \url{https://huggingface.co/meta-llama/Llama-3.2-3B-Instruct}.
\end{itemize}

\subsection{Training and evaluation settings}
\label{app:training-evaluation}

\paragraph{Optimization and systems.} All runs use bfloat16 and SGLang on one eight-GPU node. Qwen3-4B uses FSDP. Qwen3-30B-A3B uses FSDP for Multi-Hop QA and FSDP2 for Deep Research, with rollout tensor parallelism of four on 96GB H20 GPUs. We train for 100 policy updates with AdamW, a constant learning rate of $10^{-6}$, and weight decay $0.01$. The Adam coefficients are $(0.9,0.999)$ for Multi-Hop QA and $(0.9,0.98)$ for Deep Research. We use no auxiliary KL penalty or entropy regularization, and sample training rollouts with temperature $1.0$ and top-$p$ $1.0$. Appendix~\ref{app:seed-variability} repeats the Qwen3-4B comparison with five training seeds. TreeOPSD uses $M=4$, $N=2$, likelihood-weighted backup, target radius $\rho=0.25$, and $E=4$ sequential Forward-KL optimizer blocks, with unique-turn reconstruction.

\paragraph{Task-specific settings.} For Multi-Hop QA, collection uses 256 prompts split into four optimizer blocks of 64. Prompt and response lengths are capped at 2,048 and 6,144 tokens, retrieved observations at 16,384 characters, and each rollout at five assistant turns including the answer. Evaluation decodes one answer per question every five updates with temperature $0.6$ and top-$p$ $0.95$. For Deep Research, collection uses 128 prompts split into four blocks of 32, with prompt, response, and tool-observation limits of 4,096, 16,384, and 2,048 tokens. Training allows 60 assistant turns. BrowseComp-Plus evaluation allows 80 and greedily decodes one trajectory per question every five updates.

\paragraph{Evaluation and checkpoint selection.} We evaluate every five policy updates through step 80 on Multi-Hop QA and step 70 on BrowseComp-Plus, and report one checkpoint per run in that window. In Figure~\ref{fig:interaction-horizon}, the serialized trajectory stores two chat entries per assistant decision, so we divide both evaluation and training turn counts by two. The training-turn curve averages the rollouts collected at each checkpoint. Final Deep Research evaluation uses that checkpoint with one rollout per question, exact match against the short reference, at most 200 assistant turns, and a 131,072-token response budget. The reported average weights the four datasets equally.

\subsection{Baselines and rollout-budget control}
\label{app:baseline-budget}

\paragraph{Controlled methods.} ReAct \citep{yao2022react} evaluates the shared initialization without parameter updates. GRPO \citep{shao2024deepseekmath} standardizes terminal rewards over complete rollouts from each prompt. GSPO \citep{zheng2025gspo} replaces token-level importance ratios with a sequence-level ratio while retaining group-relative terminal advantages. TreePO uses the same initialize-then-expand trees, likelihood-weighted credits $V_i$, unique-turn reconstruction, prompt normalization, and optimizer blocks as TreeOPSD, but applies the fixed ascent coefficient $V_i/L_x$ throughout optimization instead of matching a finite posterior target. Every trainable method applies four actor optimizer blocks per collected prompt batch.

\paragraph{Rollout budget.} GRPO and GSPO use $G=8$ rollouts per prompt. TreePO and TreeOPSD begin with $M=4$ complete trajectories, select $N=2$ branch-anchor occurrences from each, and sample one fresh terminal suffix per occurrence. Keeping the roots gives $M(N+1)=12$ terminal leaves. A continuation randomly starts from an intermediate prefix, so its expected generation length is roughly half that of a root rollout, and the expected cost $M(1+N/2)=8$ matches $G$. Appendix~\ref{app:interaction-horizon} compares the methods at equal training time.

\section{Data Examples}
\label{app:data-examples}

This section shows one training record from each setting. We print the stored system message, the tool schemas injected by the Qwen3 chat template at rollout time, the stored user prompt, and the rule-based target. The serialized RL row stores this chat prompt, the target, and per-example tool-routing metadata, but no teacher trajectory or hidden evidence.

\subsection{Multi-Hop QA (HotpotQA)}
\label{app:data-hotpotqa}

The example is the first row of the HotpotQA distractor training parquet. The system turn is the stored system string with the chat-template tool block appended. The user turn is the Search-R1 instruction followed by the question. A search call returns the top three passages from the fixed Wikipedia-2018 index with each query truncated at 512 characters.

\begin{dataexample}{Multi-Hop QA (HotpotQA) Example}
\textbf{System prompt:}
You are a helpful and harmless assistant.

\# Tools

You may call one or more functions to assist with the user query.

You are provided with function signatures within \texttt{<tools></tools>} XML tags:
\begin{verbatim}
<tools>
{
  "type": "function",
  "function": {
    "name": "search",
    "description": "Searches the local wiki-18 dense retrieval
      index for passages relevant to the given queries.",
    "parameters": {
      "type": "object",
      "properties": {
        "query_list": {
          "type": "array",
          "description": "A list of fully-formed
            natural-language search queries.",
          "items": {
            "type": "string",
            "minLength": 1,
            "maxLength": 512
          },
          "minItems": 1,
          "maxItems": 4
        }
      },
      "required": ["query_list"],
      "additionalProperties": false
    },
    "strict": true
  }
}
</tools>
\end{verbatim}
For each function call, return a json object with function name and arguments within \texttt{<tool\_call></tool\_call>} XML tags:
\begin{verbatim}
<tool_call>
{"name": <function-name>, "arguments": <args-json-object>}
</tool_call>
\end{verbatim}

\textbf{Prompt:}
Answer the given question. You must conduct reasoning inside \texttt{<think>} and \texttt{</think>} first every time you get new information. After reasoning, if you find you lack some knowledge, you can call a search engine. Emit one Hermes tool call per assistant turn and no surrounding prose in this form:
\begin{verbatim}
<tool_call>
{"name":"search",
 "arguments":{"query_list":["fully formed search query"]}}
</tool_call>
\end{verbatim}
The arguments value must be a JSON object and \texttt{query\_list} must be a non-empty JSON array containing only non-empty strings. Never put a bare query, Python syntax, Markdown fences, or trailing commas inside \texttt{<tool\_call>}. The search engine will return the top searched results between \texttt{<tool\_response>} and \texttt{</tool\_response>}. You can search as many times as you want. If you find no further external knowledge needed, directly provide the answer inside \texttt{<answer>} and \texttt{</answer>}, without detailed illustrations. For example, \texttt{<answer> Beijing </answer>}. Question: Which magazine was started first Arthur's Magazine or First for Women?

\textbf{Ground-Truth Answer:}
Arthur's Magazine
\end{dataexample}

\subsection{Deep Research (OpenResearcher)}
\label{app:data-openresearcher}

RL construction reads one OpenResearcher record per unique question and discards the teacher messages~\citep{li2026openresearcher}. All reported runs present only the question as the user turn, together with the system string below. The training row routes \texttt{browser.search}, \texttt{browser.open}, and \texttt{browser.find} to the fixed OpenResearcher corpus. Evaluation rows keep the same interface and select the fixed BrowseComp-Plus corpus or the Serper backend.

\begin{dataexample}{Deep Research (OpenResearcher) Example}
\textbf{System prompt:}
You are a helpful assistant and harmless assistant.

You will be able to use a set of browsering tools to answer user queries.

Tool for browsing.
The \texttt{`cursor`} appears in brackets before each browsing display: \texttt{`[\{cursor\}]`}.
Cite information from the tool using the following format:
\texttt{`[\{cursor\}$\dagger$L\{line\_start\}(-L\{line\_end\})?]`}, for example: \texttt{`[6$\dagger$L9-L11]`} or \texttt{`[8$\dagger$L3]`}.
Do not quote more than 10 words directly from the tool output.
\texttt{sources=web}

When you have enough evidence, output your final answer in the exact form \texttt{`<answer>YOUR\_ANSWER</answer>`} on its own line. Only the concise answer string goes inside; no citations, no explanations, no extra text after \texttt{`</answer>`}.
Example final line:\texttt{~~`<answer>Paris</answer>`}

\# Tools

You may call one or more functions to assist with the user query.

You are provided with function signatures within \texttt{<tools></tools>} XML tags:
\begin{verbatim}
<tools>
{
  "type": "function",
  "function": {
    "name": "browser.search",
    "description": "Searches for information related to a
      query and displays top N results with titles, URLs,
      and summaries.",
    "parameters": {
      "type": "object",
      "properties": {
        "query": {
          "type": "string",
          "description": "The search query string."
        },
        "topn": {
          "type": "integer",
          "description": "Number of results to display; default 10."
        }
      },
      "required": ["query"]
    }
  }
}
{
  "type": "function",
  "function": {
    "name": "browser.open",
    "description": "Opens a zero-based result id from the
      most recent browser.search and displays a
      line-numbered viewport. Use the number inside
      [id title]; do not use the [cursor] page number
      or an L line number as id.",
    "parameters": {
      "type": "object",
      "properties": {
        "id": {
          "type": "integer",
          "description": "Zero-based result id shown inside
            [id title], not the leading L line number or
            [cursor] page number. Default -1 reopens the
            current page."
        },
        "loc": {
          "type": "integer",
          "description": "Starting line; default -1."
        },
        "num_lines": {
          "type": "integer",
          "description": "Number of lines; default -1."
        }
      },
      "required": []
    }
  }
}
{
  "type": "function",
  "function": {
    "name": "browser.find",
    "description": "Finds exact case-insensitive matches
      in an opened page.",
    "parameters": {
      "type": "object",
      "properties": {
        "pattern": {
          "type": "string",
          "description": "The exact text pattern to search for."
        },
        "cursor": {
          "type": "integer",
          "description": "Page cursor; default -1 means
            the most recent page."
        }
      },
      "required": ["pattern"]
    }
  }
}
</tools>
\end{verbatim}
For each function call, return a json object with function name and arguments within \texttt{<tool\_call></tool\_call>} XML tags:
\begin{verbatim}
<tool_call>
{"name": <function-name>, "arguments": <args-json-object>}
</tool_call>
\end{verbatim}

\textbf{Prompt:}
Determine if Neil Patrick Harris has narrated the audiobook version of 'Henry Huggins'.

\textbf{Ground-Truth Answer:}
Yes, Neil Patrick Harris narrated the audiobook version of *The Henry Huggins Audio Collection*.
\end{dataexample}

\end{document}